\documentclass{article}

\newcommand*\samethanks[1][\value{footnote}]{\footnotemark[#1]}

\usepackage{amsmath,amsfonts,bm}

\def\eqref#1{equation~\ref{#1}}

\def\1{\bm{1}}

\DeclareMathAlphabet{\mathsfit}{\encodingdefault}{\sfdefault}{m}{sl}
\SetMathAlphabet{\mathsfit}{bold}{\encodingdefault}{\sfdefault}{bx}{n}

\usepackage{PRIMEarxiv}

\usepackage{hyperref}
\usepackage{url}
\usepackage{bm}

\usepackage[utf8]{inputenc} 
\usepackage{natbib}
\setcitestyle{numbers,square}

\usepackage[T1]{fontenc}    
\usepackage{hyperref}       
\usepackage{url}            
\usepackage{booktabs}       
\usepackage{amsfonts}       
\usepackage{nicefrac}       
\usepackage{microtype}      
\usepackage{lipsum}
\usepackage{subfiles}
\usepackage{fancyhdr}       
\usepackage{graphicx}       
\usepackage{amsthm}
\usepackage{amssymb}
\usepackage[x11names]{xcolor}         
\usepackage[english]{babel}
\newtheorem{theorem}{Theorem}
\newtheorem{definition}{Definition}
\usepackage[english]{babel}
\usepackage{caption}
\usepackage{subcaption}
\usepackage{rotating}

\usepackage{algorithm}
\usepackage{algpseudocode}
\usepackage{graphics}
\usepackage{enumitem}
\usepackage{MnSymbol,bbding,pifont}
\usepackage[x11names]{xcolor}         

\newcommand{\squishlist}{
\begin{list}{{{\small{$\bullet$}}}}
{\setlength{\itemsep}{3pt}      \setlength{\parsep}{1pt}
\setlength{\topsep}{1pt}       \setlength{\partopsep}{0pt}
\setlength{\leftmargin}{1em} \setlength{\labelwidth}{1em}
\setlength{\labelsep}{0.5em} } }
\newcommand{\squishend}{  \end{list}  }
\newcommand{\btheta}{\boldsymbol{\theta}}

\title{Label Smoothing Improves Machine Unlearning
}

\author{
  Zonglin Di, Zhaowei Zhu \\
  University of California, Santa Cruz\\
  \texttt{\{zdi, zzhu61\}@ucsc.edu} \\
  \And
  Jinghan Jia\thanks{For equal contribution}, Jiancheng Liu\samethanks \\
  Michigan State University \\
  \texttt{\{jiajingh, liujia45\}@msu.edu} \\
  \And
  Zafar Takhirov, Bo Jiang \\
  Tiktok \\
  \texttt{\{bjiang518, z.tahirov\}@gmail.com} \\
  \And
  Yuanshun Yao \\
  Bytedance Research \\
  \texttt{kevinyaowork@gmail.com} \\
  \And
  Sijia Liu \\
  Michigan State University \\
  \texttt{liusiji5@msu.edu} \\
  \And
  Yang Liu \\
  University of California, Santa Cruz \\
  \texttt{yangliu@ucsc.edu} \\
}

\begin{document}
\maketitle

\begin{abstract}
The objective of machine unlearning (MU) is to eliminate previously learned data from a model. 
However, it is challenging to strike a balance between computation cost and performance when using existing MU techniques. 
Taking inspiration from the influence of label smoothing on model confidence and differential privacy, we propose a simple gradient-based MU approach that uses an inverse process of label smoothing. 
This work introduces UGradSL, a simple, plug-and-play MU approach that uses smoothed labels. 
We provide theoretical analyses demonstrating why properly introducing label smoothing improves MU performance. 
We conducted extensive experiments on six datasets of various sizes and different modalities, demonstrating the effectiveness and robustness of our proposed method. 
The consistent improvement in MU performance is only at a marginal cost of additional computations.  
For instance, UGradSL improves over the gradient ascent MU baseline by 66\% unlearning accuracy without sacrificing unlearning efficiency. 
\end{abstract}


\section{Introduction}

Building a reliable ML model has become an important topic in this community. Machine unlearning (MU) is a task requiring removing the learned data points from the model. The concept and the technology of MU enable researchers to delete sensitive or improper data in the training set to improve fairness, robustness, and privacy and get a better ML model for product usage \citep{chen2021machine, sekhari2021remember}.
Retraining from scratch ({Retrain}) is a straightforward method when we want to remove the data from the model; yet it incurs prohibitive computation costs for large models due to computing resource constraints. Therefore, an efficient and effective MU method is desired.


The most straightforward MU approach should be retraining-based method \citep{bourtoule2021machine}, meaning that we retrain the model from scratch without using the data to be forgotten. The method can guarantee privacy protection but the computational cost is intensive. 
Most existing works \citep{koh2017understanding, golatkar2020eternal, warnecke2021machine, graves2021amnesiac, thudi2021unrolling, izzo2021approximate, becker2022evaluating, jia2023model} focus on \textit{approximate MU} to achieve a balance between unlearning efficacy and computational complexity, making them more suitable for real-world applications, meaning that make the model unlearn the forgetting dataset without retraining the model.

We desire an approach that enjoys both high performance and fast speed. Since MU can be viewed as the inverse process of  ML, we are motivated to think it would be a natural and efficient way to develop an unlearning process that imitates the reverse of gradient descent. Indeed, gradient ascent (GA)  \citep{thudi2021unrolling} is one of the MU methods but unfortunately, it does not fully achieve the potential of this idea. One of the primary reasons is that once the model completes training, the gradient of well-memorized data that was learned during the process is diminishing (close to 0 loss) and therefore the effect of GA is rather limited.

Our approach is inspired by the celebrated idea of label smoothing \citep{szegedy2016rethinking}. In the forward problem (gradient descent), the smoothed label proves to be able to improve the model's generalization power. 
{In our setting, we treat the smoothed label term as the regularization in the loss function, making the unlearning more controllable.}
Specifically, we show that GA with a ``negative" label smoothing process (which effectively results in a standard label smoothing term in a descending fashion) can quickly improve the model's deniability in the forgetting dataset, which is exactly the goal of MU. We name our approach \textit{UGradSL}, unlearning using gradient-based smoothed labels. 

Our approach is a plug-and-play method that can improve the gradient-based MU performance consistently and does not hurt the performance of the remaining dataset and the testing dataset in a gradient-mixed way.
At the same time, we provide a theoretical analysis of the benefits of our approach for the MU task. 
The core contributions of this paper are summarized as follows:
\vspace{-2pt}
\squishlist
    \item We propose a lightweight tool to improve MU by joining the label smoothing and gradient ascent.
    \item We theoretically analyze the role of gradient ascent in MU and how negative label smoothing is able to boost MU performance.
    \item Extensive experiments in six datasets in different modalities and several unlearning paradigms regarding different MU metrics show the robustness and generalization of our method.
\squishend
\section{Related Work}

\textbf{Machine Unlearning} (MU) was developed to address information leakage concerns related to private data after the completion of model training \citep{cao2015towards, bourtoule2021machine, nguyen2022survey}, gained prominence with the advent of privacy-focused legislation \citep{hoofnagle2019european,pardau2018california}. One direct unlearning method involves retraining the model from scratch after removing the forgetting data from the original training set. It is computationally inefficient, prompting researchers to focus on developing approximate but much faster unlearning techniques \citep{becker2022evaluating, golatkar2020eternal, warnecke2021machine, graves2021amnesiac, thudi2021unrolling, izzo2021approximate, jia2023model}. Beyond unlearning methods, other research efforts aim to create probabilistic unlearning concepts \citep{ginart2019making, guo2019certified, neel2021descent, ullah2021machine, sekhari2021remember} and facilitate unlearning with provable error guarantees, particularly in the context of differential privacy (DP) \citep{dwork2006our, ji2014differential, hall2012differential}.
However, it typically necessitates stringent model and algorithmic assumptions, potentially compromising effectiveness against practical adversaries, such as membership inference attacks \citep{graves2021amnesiac, thudi2021unrolling}. Additionally, the interest in MU has expanded to encompass various learning tasks and paradigms
\citep{wang2022federated, liu2022right, chen2022graph, chien2022certified, marchant2022hard, di2022hidden}. These applications demonstrate the growing importance of MU techniques in safeguarding privacy. 

\textbf{Label Smoothing} (LS) or positive label smoothing (PLS) \citep{szegedy2016rethinking} is a commonly used regularization method to improve the model performance.
Standard training with one-hot labels will lead to overfitting easily. 
Empirical studies have shown the effectiveness of LS in noisy label \citep{szegedy2016rethinking, pereyra2017regularizing, vaswani2017attention, chorowski2016towards}.
In addition, LS shows its capability to reduce overfitting, improve generalization, etc.
LS can also improve the model calibration \citep{muller2019does}.
However, most of the work about LS is PLS. \citep{wei2021smooth} first proposes the concept of negative label smoothing and shows there is a wider feasible domain for the smoothing rate when the rate is negative, expanding the usage of LS. 

\textbf{Influence Function} is a classic statistical method to track the impact of one training sample. 
\cite{koh2017understanding} uses a second-order optimization approximation to evaluate the impact of a training sample.
Additionally, it can also be used to identify the importance of the training groups \citep{basu2020second, koh2019accuracy}.
The influence function is widely used in many machine-learning tasks. 
such as data bias solution \citep{brunet2019understanding, kong2021resolving}, fairness \citep{sattigeri2022fair, wang2022understanding}, security \citep{liu2022backdoor}, transfer learning \citep{jain2022data}, out-of-distribution generalization \citep{ye2021out}, etc. 
The approach also plays an important role as the algorithm backbone in the MU tasks \citep{jia2023model, warnecke2021machine, izzo2021approximate}. 

\textbf{Differential Privacy} (DP) is a mathematical framework designed to quantify and mitigate privacy risks in machine learning models. It ensures that the inclusion or exclusion of a single data point in a dataset does not significantly affect the model's output, thus protecting individual data points from being inferred by adversaries \cite{dwork2006our}. In machine learning, DP mechanisms such as noise addition and gradient clipping are employed during the training process to provide formal privacy guarantees while maintaining model utility \cite{Abadi_2016}. These techniques help balance the trade-off between data privacy and model performance, making DP a cornerstone of privacy-preserving machine learning \cite{Shokri_2015,McMahan_2018}.


\section{Label Smoothing Enables Fast and Effective Unlearning}

This section sets up the analysis and shows that properly performing 
label smoothing enables fast and effective unlearning. The key ingredients of our approach are gradient ascent (GA) and label smoothing (LS). We start with understanding how GA helps with unlearning and then move on to show the power of LS. At the end of the section, we formally present our algorithm. 
\subsection{Preliminary}

\textbf{Machine Unlearning}
Consider a $K$-class classification problem on the training data distribution $\mathcal{D}_{tr} = (\mathcal{X} \times \mathcal{Y})$, where $\mathcal{X}$ and $\mathcal{Y}$ are feature and label space, respectively. Due to some privacy regulations, there exists a forgetting data distribution $\mathcal{D}_f$ that the model needs to unlearn.
We denote by $\btheta_{tr}$ the original model trained on $\mathcal{D}_{tr}$ and $\btheta_{u}$ the model without the influence of $\mathcal{D}_f$.
The goal of machine unlearning (MU) is how to generate $\btheta_u$ from $\btheta_{tr}$.

\textbf{Label Smoothing}
In a $K$-class classification task, let $\boldsymbol{y}_i$ denote the one-hot encoded vector form of $y_i \in \mathcal{Y}$.
Similar to \cite{wei2021smooth}, we unify positive label smoothing (PLS) and negative label smoothing (NLS) into generalized label smoothing (GLS). The random variable of smoothed label $\boldsymbol{y}^{GLS, \alpha}_i$ with smooth rate $\alpha \in (-\infty, 1]$ is $\boldsymbol{y}_i^{\text{GLS}, \alpha} = (1-\alpha) \cdot \boldsymbol{y}_i + \frac{\alpha}{K} \cdot \boldsymbol{1} = [\frac{\alpha}{K}, \cdots, \frac{\alpha}{K}, (1+\frac{1-K}{K}\alpha), \frac{\alpha}{K}, \cdots, \frac{\alpha}{K}],$
where $(1+\frac{1-K}{K}\alpha)$ is the $y_i$th element in the encoded label vector. When $\alpha<0$,  GLS becomes NLS. 

\subsection{Gradient Ascent Can Help Gradient-Based Machine Unlearning}
\label{sec:ga_help}
We discuss three sets of model parameters in the MU problem: 1) $\btheta_{tr}^*$, the optimal parameters trained from $D_{tr} \sim \mathcal{D}_{tr}$ 
, 2) $\btheta_{r}^*$, the optimal parameters trained from $D_{r} \sim \mathcal{D}_r$, such that $D_{r} = D_{tr}\setminus D_{f}$ and 3) $\btheta_{f}^*$, the optimal parameters unlearnt using gradient ascent (GA) on $D_f\sim \mathcal{D}_f$. 
Note $\btheta_r^*$ can be viewed as the \textit{exact} MU model. 
The definitions of $\btheta_{tr}^{*}$ and $\btheta_{r}^{*}$ are similar to Equation \ref{equ:min_erm} and by using the influence function, $\btheta_{f}^{*}$ is 
\[
    \btheta_f^{*} = \arg \min _{\btheta} R_f(\btheta) = \arg \min _{\btheta} \{R_{tr}(\btheta)+\varepsilon \sum_{z^{f} \in D_{f}}\ell(h_{\btheta},z^{f})\}
\]
where $R_{tr}(\btheta)=\sum_{z^{tr} \in D_{tr}} \ell(h_{\btheta}, z^{tr})$ and $R_{f}(\btheta)=\sum_{z^{f} \in D_{f}} \ell(h_{\btheta, z^{f}})$ are the empirical risk on $D_{tr}$ and $D_{f}$, respectively. $\varepsilon$ is the weight of $D_f$ compared with $D_{tr}$.
The optimal parameter can be found when the gradient is 0:
\begin{equation}
\label{equ:unlearning_opt}
    \nabla_{\btheta} R_{f}(\btheta_{f}^{*}) = \nabla_{\btheta} R_{tr}(\btheta_{f}^{*}) + \varepsilon \sum_{z^{f} \in D_f}\nabla_{\btheta}\ell(h_{\btheta_{f}^{*}}, z^{f}) = 0.
\end{equation}

Expanding Eq.~(\ref{equ:unlearning_opt}) at $\btheta = \btheta_{tr}^{*}$ using the Taylor series, we have
\begin{equation}
\label{equ:f_tr_para}
    \btheta^{*}_{f} - \btheta^{*}_{tr}  \approx -\left[\sum_{z^{tr} \in D_{tr}} \nabla_{\btheta}^{2} \ell(h_{\btheta_{tr}^{*}}, z^{tr}) + \varepsilon \sum_{z^{f} \in D_{f}}\nabla_{\btheta}^2 \ell(h_{\btheta_{tr}^{*}}, z^{f}) \right]^{-1}
    \left( \varepsilon \sum_{z^{f} \in D_{f}} \nabla_{\btheta} \ell(h_{\btheta_{tr}^{*}}, z^{f})\right).
\end{equation}



Similarly, we can expand $R_{tr}(\btheta_{tr}^*)$ at $\btheta = \btheta_r^{*}$ and derive $\btheta_{r}^* - \btheta_{tr}^*$ as 
\[
    \btheta^{*}_{r} - \btheta^{*}_{tr} \approx \left(\sum_{z^{tr} \in D_{tr}} \nabla^2_{\btheta} \ell(h_{\btheta_{r}^{*}},z^{tr})\right)^{-1}\left(\sum_{z^{tr} \in 
 D_{tr}} \nabla_{\btheta} \ell(h_{\btheta_{r}^{*}}, z^{tr})\right).
\]

We ignore the average operation in the original definition of the influence function for computation convenience because the size of $D_{tr}$ or $D_f$ are fixed. 
For GA, let $\varepsilon = -1$ in Equation \ref{equ:f_tr_para} and we have
\begin{equation}
\label{equ:param_distance}
    \btheta_{r}^{*} - \btheta_{f}^{*} \approx \btheta_{r}^{*} - \btheta_{tr}^{*} - (\btheta^{*}_{f} - \btheta^{*}_{tr}) = {\Delta \btheta_{r}}  - {\Delta \btheta_{f}}
\end{equation}
$(-\Delta \btheta_{r})$ represents the learning gap from $\btheta_r^{*}$ to $\btheta_{tr}^{*}$ while 
vector $\Delta \btheta_{f}$ represents how much the model unlearns (backtracked progress) between $\btheta_{f}^*$ and $\btheta_{tr}^*$.
The details of $\Delta \btheta_r$ and $\Delta \btheta_f$ are given in Equation \ref{equ:r_k_diff} in the Appendix. Ideally, when $\Delta \btheta_r$ and $\Delta \btheta_f$ are exactly the same vectors, GA can lead the model to the optimal retrained model since we have $\btheta_{r}^{*} = \btheta_{f}^{*}$. However, this condition is hard to satisfy in practice. Thus, GA cannot always help MU. We summarize it in Theorem~\ref{the:GA_helps} and the proof is given in Section \ref{sec:proof_GA_helps} in the Appendix.

\begin{theorem}
\label{the:GA_helps}
Given the approximation in Equation~\ref{equ:param_distance}, GA achieve exact MU if and only if 
\[
\sum_{z^{f} \in D_{f}} \nabla_{\btheta} \ell(h_{\btheta_{r}^{*}}, z^{f}) =  - \bm H(\btheta_{r}^{*}, \btheta_{tr}^{*}) \cdot \sum_{z^{f} \in D_{f}}\nabla_{\btheta} \ell(h_{\btheta^{*}_{tr}}, z^{f}),
\]
where ${\bm H(\btheta_{r}^{*}, \btheta_{tr}^{*}) = \left(\sum_{z^{tr} \in D_{tr}} \nabla_{\btheta}^2 \ell(h_{\btheta_{r}^{*}}, z^{tr})\right) \left(\sum_{z^{r} \in D_r}\nabla^2_{\btheta} \ell(h_{\btheta_{tr}^{*}}, z^{r})\right)^{-1}}$. Otherwise, there exist $\btheta_{r}^{*}, \btheta_{tr}^{*}$ such that GA can not help MU, i.e., 
$
 \|\btheta_r^{*} - \btheta_f^{*}\| > \|\btheta_r^* - \btheta_{tr}^*\|.
$

\end{theorem}




\subsection{Label Smoothing Improves MU}
\label{sec:nls_MU}

In practice, we cannot guarantee that GA always helps MU as shown in Theorem~\ref{the:GA_helps}. To alleviate the possible undesired effect of GA, we propose to use label smoothing as a plug-in module.
Consider the cross-entropy loss as an example.
For GLS, the loss is calculated as
\begin{equation}
\label{equ:gls}
    \ell(h_{\btheta}, z^{\text{GLS}, \alpha}) = \left(1+\frac{1-K}{K}\alpha\right)\cdot \ell(h_{\btheta}, (x,y)) + \frac{\alpha}{K}\sum_{y^{\prime}\in \mathcal{Y} \backslash y} \ell(h_{\btheta}, (x,y^{\prime})),
\end{equation}
where we use notations $\ell(h_{\btheta}, (x,y)):=\ell(h_{\btheta}, z)$ to specify the loss of an example $z=(x, y)$ in the dataset and $\ell(h_{\btheta}, (x,y^{\prime}))$ to denote the loss of an example when its label is replaced with $y^{\prime}$. 
Intuitively, Term $\sum_{y^{\prime} \in \mathcal{Y} \backslash y} \ell(h_{\btheta}, (x, y^{\prime}))$ in Equation \ref{equ:gls} 
leads to a state where the model makes \textit{wrong predictions on data in the forgetting dataset} with equally low confidence \citep{wei2021smooth,lukasik2020does}. 

With smoothed label given in Equation \ref{equ:gls}, we show that there exists a vector $\Delta \btheta_n$ such that equation \ref{equ:param_distance} can be written as 
\begin{equation}
    \btheta_{r}^{*} - \btheta_{f,\text{LS}}^{*} \approx \Delta \btheta_{r} - \Delta \btheta_{f} + \frac{1-K}{K}\alpha \cdot (\Delta \btheta_n-\Delta \btheta_{f}),
\end{equation}
We leave the detailed form of $\Delta \btheta_n$ to 
after Equation~\ref{equ:param_distance_NLS} in the Appendix. But intuitively, $\Delta \btheta_n$ captures the gradient influence of the smoothed non-target label on the weight. We show the effect of NLS ($\alpha < 0$) in Theorem~\ref{the:negative_is_pos} below and its proof is given in Section \ref{sec:negative_is_pos} in the Appendix.
\begin{theorem}
\label{the:negative_is_pos}
Given the approximation in Equation~\ref{equ:param_distance} and $\langle \Delta \btheta_{r} - \Delta \btheta_{f}, \Delta \btheta_n-\Delta \btheta_{f}\rangle \le 0$, there exists an $\alpha < 0$ such that GLS improves GA in unlearning, i.e., $\|\btheta_{r}^{*} - \btheta_{f, \text{GLS}}^{*}\| < \|\btheta_{r}^{*} - \btheta_{f}^{*}\|$,
where $\btheta^*_{f, \text{GLS}}$ is the optimal parameters unlearned using GA and GLS, and $\langle \cdot, \cdot \rangle$ the inner product of two vectors.
\end{theorem}

Now we explain the above theorem intuitively. Vector $ \Delta \btheta_{f} - \Delta \btheta_{r}$ is the resultant of Newton's direction of learning and unlearning. Vector $\Delta \btheta_f-\Delta \btheta_{n}$ is resultant of Newton's direction of learning non-target labels and unlearning the target label.
When the condition $\langle \Delta \btheta_{r} - \Delta \btheta_{f}, \Delta \btheta_n-\Delta \btheta_{f}\rangle \le 0$ holds, $\Delta \btheta_n-\Delta \btheta_{f}$ captures the effects of the smoothing term in the unlearning process. If we assume that the exact MU model is able to fully unlearn an example, vector $\Delta \btheta_n$ contributes a direction that pushes the model closer to the exact MU state by leading the model to give the wrong prediction. 
The illustration of $\langle \Delta \btheta_{r} - \Delta \btheta_{f}, \Delta \btheta_n-\Delta \btheta_{f}\rangle$ is shown in Figure \ref{fig:grad_distribution} in the Appendix.

\paragraph{Label smoothing helps and local differential privacy} When $\alpha < 0$, the smoothing term will incur a positive effect in the gradient ascent step. Label smoothing can also be viewed through the lens of privacy protection. This interpretation stems from the fact that label smoothing reduces the likelihood of a specific label, thereby allowing it to better blend in with other candidate labels.
Particularly, we consider a local differential privacy (LDP) guarantee for labels as follows.
\begin{definition}[Label-LDP]\label{def:label-ldp}
A privacy protection mechanism $\mathcal{M}$ satisfies $\epsilon$-Label-LDP, if for any labels $y,y', y^{\texttt{pred}}\in \mathcal{Y}$,
\begin{equation*}
    \frac{\mathbb P\left(\mathcal{M}(y) = y^{\texttt{pred}}\right)}{\mathbb P\left(\mathcal{M}(y') = y^{\texttt{pred}}\right)}\le e^{\epsilon}.
\end{equation*}
\end{definition}
The operational meaning of $\mathcal{M}$ is to guarantee any two labels $y$ and $y'$ in the label space, after privatization, have a similar likelihood to become any $y^{\texttt{pred}}$ in the label space. That is, the prediction on the forgetting dataset should be similar no matter what the ground-truth label is. 
The similarity is measured by the privacy budget $\epsilon\in[0,+\infty)$. Smaller $\epsilon$ implies strong indistinguishability between $y$ and $y'$, and hence, stricter privacy. 



Recall $R_{tr}(\btheta)=\sum_{z^{tr} \in D_{tr}} \ell(h_{\btheta}, z^{tr})$. Denote by $R^{\text{NLS}}_{f}(\btheta; \alpha)=\sum_{z^{\text{LS},\alpha} \in D_{f}} \ell(h_{\btheta}, z^{\text{NLS},\alpha}), \alpha < 0$ the empirical risk of forgetting data with NLS. After MU with label smoothing on $D_f$ by gradient ascent, the resulting model can be seen as minimizing the risk $\gamma_1 \cdot R_{tr}(\btheta) - \gamma_2 \cdot R^{\text{NLS}}_{f}(\btheta; \alpha)$, which is a weighted combination of the risk from two phases: 1) machine learning on $D_{tr}$ with weight $\gamma_1>0$ and 2) machine unlearning on $D_{f}$ with weight $\gamma_2>0$. 
By analyzing the risk, we have the following theorem to show NLS in MU induces $\epsilon$-Label-LDP for the forgetting data. 
\begin{theorem}
\label{thm:erm}
Suppose $\gamma_1 - \gamma_2(1+\frac{1-K}{K}\alpha)  > 0$. MU using GA+LS 
achieves $\epsilon$-Label-LDP on $D_f$ where
    \begin{equation*}
        \epsilon = \left|\log\left(  \frac{K}{\alpha}\left(1-\frac{\gamma_1}{\gamma_2}\right) + 1 - K \right) \right |, ~ \alpha < 0.
    \end{equation*}
\end{theorem}
Intuitively, when $\alpha$ is more negative, the privacy of the labels in the forgetting dataset is better. When $\alpha \rightarrow (1-\gamma_1/\gamma_2)$, we have $\epsilon \rightarrow 0$, indicating the best label-LDP result, which is the goal of MU. The theorem also warns that $\alpha$ cannot be arbitrarily negative. 

\subsection{UGradSL: A Plug-and-Play and Gradient-Mixed MU Method}\label{sec:framework}

Compared with retraining, Fine-Tune (FT) and GA are much more efficient as illustrated in the Experiment part in Section \ref{sec:exp_res} with comparable or better MU performance. FT and GA focus on different perspectives of MU. FT is to transfer the knowledge of the model from $D_{tr}$ to $D_{r}$ using gradient descent (GD) while GA is to remove the knowledge of $D_{f}$ from the model. 
Due to the flexibility of label smoothing, our method is suitable for the gradient-based methods including FT and GA, making our method a plug-and-play algorithm.
UGradSL is based on GA while UGradSL+ is on FT. 
Compared with UGradSL, UGradSL+ will lead to a more comprehensive result but with a larger computation cost. 
The algorithm is presented in Algorithm \ref{alg:cap}. 
The smooth rate $\alpha_i$ for each point $z_i^f$ in $D_f$ is not even in $D_f$. 
It depends on its distance $d(z_i^r, z_i^f) \in [0, 1]$ for each $(z_i^r, z_i^f)$ pair.
The intuition is that if an instance $z_i^f$ resides in a dense neighborhood, its inherent deniability is higher and therefore the requirement for ``forgetting" is lesser and should be reflected through a smaller $\alpha_i$. 
We leave the details of the implementation in Section \ref{sec:algo_details} in the Appendix. 





Assuming the amount of retained data is significantly larger than the amount of data to be forgotten ($|D_r| > |D_f|$), $D_{f}$ will be iterated several times when $D_r$ is fully iterated once.
We calculate the loss using a gradient-mixed method as:
\begin{equation}
\label{equ:loss_calculation}
    L(h_{\btheta}, B_{f}^{\text{GLS}, \alpha}, B_{r}, p) = p \cdot \sum_{z^r \in B_r} \ell(h_{\btheta}, z^r) - (1-p) \cdot \sum_{z_i^{f, \text{GLS}, \alpha_i} \in B_f^{\text{GLS}, \alpha}} \ell(h_{\btheta}, z_i^{f, \text{GLS}, \alpha_i})
\end{equation}
where $p \in [0, 1]$ is used to balance GD and GA and the minus sign stands for the GA. $h_{\btheta}$ is updated according to $L$ (Line 9).
UGradSL is similar to UGradSL+ and the dataset used is given in bracket in Algorithm \ref{alg:cap}.
The difference between UGradSL and UGradSL+ is the convergence standard.
UGradSL is based on the convergence of $D_f$ while UGradSL+ is based on $D_r$.
It should be noted that the Hessian matrix in Theorem \ref{the:GA_helps} is only used in the theoretical proof. In the practical calculation, \textbf{there is no need to calculate the Hessian matrix}. 
Thus, our method does not incur substantially more computation but improves the MU performance on a large scale. We present empirical evidence in Section \ref{sec:exp_res}.
Compared with applying the label smoothing evenly, 
the proposed method takes the similarity of the data points between $D_r$ and $D_f$ into consideration and provides self-adaptive smoothed labels for individual $z^f_i$ as well as protects the LDP. 


\begin{algorithm}[t]
\caption{UGradSL+: A plug-and-play, efficient, gradient-based MU method using LS. UGradSL can be specified by imposing the  {\color{blue} dataset replacement} in the bracket. 
}
\label{alg:cap}
\begin{algorithmic}[1]
\Require A almost-converged model $h_{\hat{\btheta}_{tr}}$ trained with $D_{tr}$. The retained dataset $D_r$. The forgetting dataset $D_f$. Unlearning epochs $E$. GA ratio $p$. Distance threshold $\beta$.
\Ensure The unlearned model $h_{\btheta_f}$.
\State Set the current epoch index as $t_c \gets 1$

\While{$t_c < E$}
    \While{{$D_{r} ({\color{blue}D_{f}})$ is not fully iterated}} 
    \State {Sample a batch $B_{r}$ in $D_{r}$}
    \State {Sample a batch $B_{f}$ from $D_{f}$ where the size of $B_{f}$ is the same as that of $B_{r}$}
    \State {Calculate the distance $d(z_i^r, z_i^f)$ for each $(z_i^r, z_i^f)$ pair where $z_i^r \in B_r$ and $z_i^f \in B_f$.}
    \State For each $z_i^f$, count the number $c_i^f$ of $z_i^r$ whose $d(z_i^r, z_i^f)$ < $\beta$
    \State Calculate the smooth rate $\alpha_i = c_i^f/|B_f|$ for each $z_i^f \in B_f$
    \State Update the model using $B_{r}$, $B_{f}$, $p$ and $\alpha_i$ according to Equation \ref{equ:loss_calculation}
    \EndWhile
\State $t_c \gets t_c + 1$
\EndWhile
\end{algorithmic}
\end{algorithm}

\section{Experiments and Results}
\label{sec:exp_res}

\subsection{Experiment Setup}
\textbf{Dataset and Model Selection} We validate our method using various datasets in different scales and modality, including CIFAR-10 \citep{krizhevsky2009learning}, CIFAR-100 \citep{krizhevsky2009learning}, SVHN \citep{netzer2011reading}, CelebA \citep{liu2015faceattributes}, ImageNet \citep{deng2009imagenet} and 20 Newsgroup \citep{Newsgroups20} datasets. 
For the vision and language dataset, we use ResNet-18 \citep{he2016deep} and Bert \citep{devlin2018bert} as the backbone model, respectively.
Due to the page limit, the details of the training parameter and the additional results of different models including VGG-16 \citep{simonyan2014very} and vision transformer (ViT) \citep{dosovitskiy2020image} are given in the Appendix.


\textbf{Baseline Methods}
We compare UGradSL and UGradSL+ with a series of  baseline methods, including retrain, fine-tuning (FT) \citep{warnecke2021machine, golatkar2020eternal}, gradient ascent (GA) \citep{graves2021amnesiac, thudi2021unrolling}, unlearning based on the influence function (IU) \citep{izzo2021approximate, koh2017understanding}, boundary unlearning (BU) \citep{chen2023boundary}, $\ell_1$-sparse \citep{jia2023model}, random label (RL) \citep{hayase2020selective} and SCRUB \citep{kurmanji2023towards}. Besides, there is also an unlearning paradigm called instance unlearning \citep{cha2023learning}, which is not the scope in this paper.



\textbf{Evaluation Metrics} 
The evaluation metrics we use follows \cite{jia2023model}, where we jointly consider unlearning accuracy (UA), membership inference attack (MIA), remaining accuracy (RA), testing accuracy (TA), and run-time efficiency (RTE).
UA is the ratio of incorrect prediction on $D_f$, showing the MU performance. 
TA is the accuracy used to evaluate the performance on the whole testing set $D_{te}$, except for the class-wise forgetting because the task is to forget the specific class. RA is the accuracy on $D_r$.
To evaluate the effectiveness of "forgetting", we resort to the MIA metrics described in \cite{jia2023model,fan2023salun}, i.e. accuracy of an attack model against target model $\theta_u$, such that the score is reported as true negative rate (TNR) on the forget set.
Formally, this is a global MIA score \cite{Yeom_2018}, which we rewrite as  $\mbox{MIA}_{\mbox{Score}} = 1-\Pr\left(x_f | \theta_\star\right)$, where $x_f \in D_f$ are the forget samples and $\theta_\star$ is the model under test. Note that in this setting higher $\mbox{MIA}_{\mbox{Score}}$ indicates better "forgetting" quality.



Note a \textit{tradeoff} between RA/TA and UA/$\mbox{MIA}_{\mbox{Score}}$ exists, i.e., the higher UA/$\mbox{MIA}_{\mbox{Score}}$ usually implies lower RA/TA in practice except class-wise forgetting. In our experiments, we expect RA and TA will not decrease too much (e.g., 5\%) while UA and $\mbox{MIA}_{\mbox{Score}}$ can improve consistently and more significantly. We add a discussion in Section \ref{sec:discssion}.

\begin{table*}[tb]
    \centering
    \caption{Results of class-wise forgetting in CIFAR-10 and ImageNet.
    }
    \label{tab:class_wise}
    \scalebox{0.85}{
    \begin{tabular}{c|cccc|cc|c}
    \toprule
        CIFAR-10 & UA & $\mbox{MIA}_{\mbox{Score}}$ & RA & TA & Avg. Gap ($\downarrow$) & Sum. ($\uparrow$) & RTE ($\downarrow$, min)\\
    \midrule
    Retrain & 100.00$\pm$0.00 & 100.00$\pm$0.00 & 98.19$\pm$3.14 & 94.50$\pm$0.34 & - & 392.69 & 14.92\\
        \midrule
       FT & 22.71$\pm$5.31 & 79.21$\pm$8.60 & 99.82$\pm$0.09 & 94.13$\pm$0.14 & 25.02 & 295.87 & 2.02 \\
       GA & 25.19$\pm$11.38 & 73.48$\pm$9.68 & 96.84$\pm$0.58 & 73.10$\pm$1.62 & 31.02 & 268.61 & \textbf{0.08} \\
       IU & 83.92$\pm$1.16 & 92.59$\pm$1.41 & 98.77$\pm$0.12 & 92.64$\pm$0.23 & 6.48 & 367.92 & 1.18\\
       BE & 64.93$\pm$0.01 & 98.19$\pm$0.00 & 99.47$\pm$0.00 & 94.00$\pm$0.11 & 9.67 & 356.59 & 0.20 \\
       BS & 93.69$\pm$4.32 & 99.82$\pm$0.04 & 97.69$\pm$1.29 & 92.89$\pm$1.26 & 2.15 & 384.08 & 0.29 \\
       $\ell_1$-sparse & 100.00$\pm$0.00 & 100.00$\pm$0.00 & 97.86$\pm$1.29 & 96.11$\pm$1.26 & 0.48 & 393.97 & 1.00 \\
       SCRUB & 100.00$\pm$0.00 & 100.00$\pm$0.00 & 99.93$\pm$0.01 & 95.22$\pm$0.07 & 0.62 & 395.15 & 1.09 \\
       Random Label & 99.99$\pm$0.01 & 100.00$\pm$0.00 & 100.00$\pm$0.00 & 95.50$\pm$0.11 & 0.71 & \textbf{395.49} & 1.04 \\
    \midrule
       UGradSL & 94.99$\pm$4.35 & 97.95$\pm$1.78 & 95.47$\pm$4.08 & 86.78$\pm$5.68 & 4.38 & 375.19 & 0.22 \\
       UGradSL+ & 100.00$\pm$0.00 & 100.00$\pm$0.00 & 99.26$\pm$0.01 & 94.29$\pm$0.07 & \textbf{0.32} & 392.55 & 3.07 \\
    \midrule
    ImageNet & UA & $\mbox{MIA}_{\mbox{Score}}$ & RA & TA & Avg. Gap ($\downarrow$) & Sum. ($\uparrow$) & RTE ($\downarrow$, hr)\\
    \midrule
    Retrain & 100.00$\pm$0.00 & 100.00$\pm$0.00 &71.62$\pm$0.12& 69.57$\pm$0.07 & - & 341.19 & 26.18\\ 
    \midrule
    FT  & 52.42$\pm$15.81 & 55.87$\pm$18.02 & 70.66$\pm$2.54 & 69.25$\pm$0.78 & 23.25 & 248.20 & 2.87\\
    GA & 81.23$\pm$0.69 & 83.52$\pm$2.08 & 66.00$\pm$0.03 & 64.72$\pm$0.02 & 11.43  & 295.47 & \textbf{0.01}\\
    IU  & 33.54$\pm$19.46 & 49.83$\pm$21.57 & 66.25$\pm$1.99 & 66.28$\pm$1.19 & 31.32 & 215.90 & 1.51\\
    BS & 98.85$\pm$0.50 & 0.13$\pm$0.12 & 53.35$\pm$0.16 & 56.93$\pm$0.03 & 32.98 & 209.26 & 0.37\\
    BE & 98.62$\pm$0.58 & 0.15$\pm$0.11 & 53.13$\pm$0.27 & 56.72$\pm$0.31 & 33.14 & 241.91 & 0.24\\
    $\ell1$-sparse & 100.00$\pm$0.00 & 100.00$\pm$0.00 & 39.01$\pm$1.03 & 44.62$\pm$0.91 & 14.39 & 283.63 &  0.16 \\
    SCRUB & 56.59$\pm$2.17 & 75.59$\pm$1.19 & 66.98$\pm$0.11 & 68.24$\pm$0.07 & 18.45 & 267.40 & 0.21\\
    Random Label & 100.00$\pm$0.00 & 100.00$\pm$0.00 & 62.06$\pm$4.19 & 62.93$\pm$0.45 & 16.93 & 324.99 & 1.17\\
       \midrule
       UGradSL &  100.00$\pm$0.00 & 100.00$\pm$0.00 & 76.91$\pm$1.82 & 65.94$\pm$1.35 & \textbf{2.23} & 342.85 & \textbf{0.01} \\
       UGradSL+ & 100.00$\pm$0.00 & 100.00$\pm$0.00 & 78.16$\pm$0.07 & 
        66.84$\pm$0.06 & 2.32 & \textbf{345.00} & 4.19\\
    \bottomrule
    \end{tabular}
    }
\end{table*}

\begin{table*}[t]
    \centering
    \caption{Results of random forgetting across all classes in CIFAR-100 and SVHN.}
    \label{tab:random}
    \scalebox{0.85}{%
    \begin{tabular}{c|cccc|cc|c}
    \toprule
        CIFAR-100 & UA & $\mbox{MIA}_{\mbox{Score}}$ & RA & TA & Avg. Gap ($\downarrow$) & Sum. ($\uparrow$) & RTE ($\downarrow$, min)\\
    \midrule
    Retrain & 29.47$\pm$1.59 & 53.50$\pm$1.19 & 99.98$\pm$0.01 & 70.51$\pm$1.17 & 253.46 & 25.01 & 13.08\\
    \midrule
       FT & 2.55$\pm$0.03 & 10.59$\pm$0.27 & 99.95$\pm$0.01 & 75.95$\pm$0.05 & 18.83 & 189.04 & 1.95 \\
       GA & 2.58$\pm$0.06 & 5.95$\pm$0.17 & 97.45$\pm$0.02 & 76.09$\pm$0.01 & 20.64 & 182.07 & \textbf{0.29}  \\
       IU & 15.71$\pm$5.19 & 18.69$\pm$4.12 & 84.65$\pm$5.29 & 62.20$\pm$4.17 & 18.05 & 181.25 & 1.20 \\      
       BE & 0.01$\pm$0.00 & 1.45$\pm$0.02 & 99.97$\pm$0.18 & 78.26$\pm$0.00 & 22.32 & 179.69 & 0.24 \\
       BS & 2.20$\pm$1.21 & 10.73$\pm$9.37 & 98.22$\pm$1.26 & 70.23$\pm$1.67 & 18.02 & 181.38 & 0.34\\
       $\ell_1$-sparse & 8.19$\pm$0.38 & 19.11$\pm$0.52 & 88.39$\pm$0.31 & 80.26$\pm$0.16 & 19.25 & 195.95 & 1.00 \\
       SCRUB & 0.09$\pm$0.59 & 4.01$\pm$1.25 & 99.97$\pm$0.34 & 77.45$\pm$0.26 & 21.46 & 181.52 & 1.06\\
       Random Label & 4.06$\pm$0.37 & 50.12$\pm$3.48 & 99.92$\pm$0.01 & 71.30$\pm$0.12 & 19.07 & 225.40 & 1.20 \\
    \midrule
       UGradSL & 15.10$\pm$2.76 & 34.67$\pm$0.63 & 86.69$\pm$2.41 & 59.25$\pm$2.35 & 14.44 & 195.71 & 0.55 \\
       UGradSL+ & 63.89$\pm$0.75 & 71.51$\pm$1.31 & 92.25$\pm$0.11 & 61.09$\pm$0.10 & 17.40 & \textbf{288.74} & 3.52 \\
       \midrule
       UGradSL (R) & 18.36$\pm$0.17 & 40.71$\pm$0.13 & 98.38$\pm$0.03 & 68.23$\pm$0.16 & 6.95 & 207.95 & 0.55 \\
       UGradSL+ (R) & 21.69$\pm$0.59 & 49.47$\pm$1.25 & 99.87$\pm$0.34 & 73.60$\pm$0.26 & \textbf{3.75} & 244.63 & 3.52\\
    \midrule
    20 Newsgroup & UA & $\mbox{MIA}_{\mbox{Score}}$ & RA & TA & Avg. Gap ($\downarrow$) & Sum. ($\uparrow$) & RTE ($\downarrow$, min) \\
    \midrule
    Retrain & 7.37$\pm$0.14 & 9.33$\pm$0.98 & 100.00$\pm$0.01 & 85.24$\pm$0.09 & - & 201.94 & 75.12\\
    \midrule
    FT & 2.26$\pm$1.53 & 2.70$\pm$1.60 & 98.60$\pm$0.18 & 82.20$\pm$1.12 & 4.05 & 185.76 & 1.06\\
    GA & 0.74$\pm$0.97 & 2.65$\pm$0.98 & 99.69$\pm$0.14 & 83.63$\pm$0.12 & 3.81 & 186.71 & \textbf{0.77}\\
    IU & 0.03$\pm$0.06 & 0.33$\pm$0.11 & 100.00$\pm$0.00 & 85.72$\pm$0.12 & 4.21 & 186.08 & 1.53\\
    $\ell_1$-sparse & 4.51$\pm$0.99 & 9.43$\pm$1.45 & 99.65$\pm$0.23 & 82.52$\pm$0.78 & 1.51 & 196.11 & 8.80 \\
    Random Label & 37.16$\pm$2.04 & 37.16$\pm$2.04 & 99.76$\pm$0.09 & 83.22$\pm$0.38 & 14.97 & 257.30 & 8.33\\
    \midrule 
    UGradSL & 13.00$\pm$1.17 & 44.30$\pm$1.36 & 93.46$\pm$1.01 & 73.17$\pm$0.42 & 14.80 & 223.93 & 1.50\\
    UGradSL+ & 35.53$\pm$1.53 & 46.42$\pm$0.49 & 97.15$\pm$0.88 & 78.95$\pm$1.98 & 18.60 & \textbf{258.05} & 4.73 \\
    \midrule
    UGradSL (R) & 6.16$\pm$1.17 & 7.92$\pm$1.69 & 98.09$\pm$1.73 & 81.84$\pm$2.07 & 1.98 & 194.01 & 4.73 \\
    UGradSL+ (R) & 7.50$\pm$0.48 & 9.43$\pm$0.64 & 98.54$\pm$0.56 & 81.65$\pm$1.39 & \textbf{1.32} & 197.12 & 1.50 \\
    \bottomrule
    \end{tabular}
    }
\end{table*}

\textbf{Unlearning Objective}
In addition to the individual metrics, we report two combined metrics according to different aims of MU. Specifically, we use \textit{Avg. Gap} \citep{fan2023salun} to compare the MU performance with the retraining \cite{jia2023model}, where a smaller performance gap between the unlearning and retraining indicates a better MU method. We calculate the average gap among UA, $\mbox{MIA}_{\mbox{Score}}$, RA and compare them with the retrained model. In Tables~\ref{tab:random}--\ref{tab:celebA}, we use a tag \textit{(R)} to indicate the results that are optimized for this metric.
The other of combined metric is \textit{Sum}, which is the sum of UA, TA, RA and $\mbox{MIA}_{\mbox{Score}}$. This metrics is to indicate the evaluation metric of the \textit{completely unlearning} as given in \citep{shah2023unlearning}, meaning that we want to maximize UA and $\mbox{MIA}_{\mbox{Score}}$ while maintaining the TA and RA. 

\textbf{Unlearning Paradigm}
We mainly consider three unlearning paradigms, including \textit{class-wise forgetting}, \textit{random forgetting}, and \textit{group forgetting}.
Class-wise forgetting is to unlearn the whole specific class where we remove one class in $D_r$ and the corresponding class in $D_{te}$ completely.
Random forgetting across all classes is to unlearn data points belonging to all classes.
As a special case of random forgetting, \textit{group forgetting} means that the model is trained to unlearn the group or sub-class of the corresponding super-classes. 
\begin{table*}[t]
    \centering
    \caption{The experiment results of group forgetting within one class using the CIFAR-100 dataset. The model is classify 20 super-classes and $D_f$ is one of five subclasses in one super-class.}
    \label{tab:cifar20}
   \scalebox{0.85}{%
    \begin{tabular}{c|cccc|cc|c}
    \toprule
         & UA & $\mbox{MIA}_{\mbox{Score}}$ & RA & TA & Avg. Gap ($\downarrow$) & Sum. ($\uparrow$) & RTE ($\downarrow$, min) \\
    \midrule
        Retrain & 13.33$\pm$1.64 & 28.47$\pm$0.75 & 99.94$\pm$0.01 & 81.23$\pm$0.13 & - & 229.71 & 27.35\\
    \midrule
       FT & 1.00$\pm$0.43 & 2.73$\pm$0.52 & 99.37$\pm$0.08 & 79.02$\pm$0.03 & 10.21 & 182.12 & 7.47\\
       GA & 87.93$\pm$2.92 & 88.93$\pm$2.33 & 81.46$\pm$0.77 & 64.07$\pm$0.95 & 42.68 & 322.39 & \textbf{0.11}\\
       IU & 0.00$\pm$0.00 & 2.07$\pm$1.29 & 99.95$\pm$0.01 & 80.92$\pm$0.34 & 10.01 & 18.93 & 1.10\\
       BE & 89.07$\pm$1.39 & 91.73$\pm$1.75 & 76.36$\pm$0.92 & 60.17$\pm$0.92 & 45.91 & 317.33 & 0.33 \\
       BS & 88.60$\pm$1.13 & 90.67$\pm$1.18 & 76.70$\pm$1.08 & 60.41$\pm$1.17 & 45.38 & 316.37 & 0.29\\
       $\ell_1$-sparse & 0.13$\pm$0.09 & 2.27$\pm$0.57 & 99.57$\pm$0.04 & 80.44$\pm$0.08 & 10.14 & 182.41 & 0.38 \\
       SCRUB & 0.00$\pm$0.00 & 1.13$\pm$0.34 & 99.93$\pm$0.01 & 81.05$\pm$0.20 & 10.21 & 182.12 & 0.30\\
       Random Label & 56.93$\pm$3.24 & 98.60$\pm$0.29 & 99.92$\pm$0.01 & 80.28$\pm$0.05 & 28.67 & 335.74 & 0.37\\
    \midrule
       UGradSL & 79.56$\pm$0.34 & 90.44$\pm$1.99 & 98.16$\pm$4.46 & 80.06$\pm$0.90 & 32.79 & \textbf{348.22} & 0.13 \\
        UGradSL+ & 81.11$\pm$0.74 & 86.96$\pm$1.89 & 95.69$\pm$0.82 & 78.02$\pm$1.02 & 33.43 & 341.78 & 8.12\\
    \midrule
        UGradSL (R) & 22.87$\pm$0.90 & 38.93$\pm$1.57 & 97.20$\pm$0.19 & 75.84$\pm$0.16 & \textbf{7.03} & 234.84& 0.13 \\ 
        UGradSL+ (R) & 78.44$\pm$1.19 & 88.67$\pm$0.35 & 97.93$\pm$0.71 & 79.77$\pm$0.58 & 2.40 & 344.81 & 8.12\\
    \bottomrule
    \end{tabular}
    }
\end{table*}

\begin{table*}[t]
    \centering
    \caption{The experiment results of group forgetting in CelebA. The model is to classify whether the person is smile or not and $D_f$ is selected according to the identities.}
    \label{tab:celebA}
    \scalebox{0.85}{%
    \begin{tabular}{c|cccc|cc|c}
    \toprule
         & UA & $\mbox{MIA}_{\mbox{Score}}$ & RA & TA & Avg. Gap ($\downarrow$) & Sum. ($\uparrow$) & RTE ($\downarrow$, min) \\
    \midrule
        Retrain & 6.74$\pm$0.26 & 9.77$\pm$1.49 & 94.38$\pm$0.49 & 91.78$\pm$0.33 & 63.33 & 202.67 & 258.69\\
    \midrule
       FT & 5.36$\pm$0.17 & 5.87$\pm$0.11 & 93.91$\pm$0.04 & 93.18$\pm$0.03 & 1.79 & 198.32 & 25.94 \\
       GA & 6.00$\pm$0.16 & 5.76$\pm$0.14 & 92.86$\pm$0.13 & 92.52$\pm$0.08 & 1.75 & 197.14 & \textbf{1.20}\\
       IU & 5.90$\pm$0.11 & 4.91$\pm$0.30 & 93.05$\pm$0.01 & 92.62$\pm$0.01 & 1.97 & 196.48 & 219.77 \\
       BE & 11.50$\pm$0.80 & 48.41$\pm$8.86 & 88.37$\pm$0.81 & 88.07$\pm$0.81 & 13.28 & \textbf{236.35} & 48.91\\
       BS & 8.95$\pm$5.11 & 27.35$\pm$30.20 & 91.00$\pm$5.22 & 90.63$\pm$5.65 & 6.08 & 217.93 & 50.99\\
       $\ell_1$-sparse & 9.46$\pm$1.82 & 36.91$\pm$30.96 & 90.52$\pm$1.75 & 90.35$\pm$1.77 & 8.79 & 227.24 & 37.49\\
       SCRUB & 8.78$\pm$0.77 & 13.37$\pm$5.22 & 91.21$\pm$0.86 & 90.65$\pm$0.86 & 2.49 & 204.01 & 70.13\\
       Random Label & 8.31$\pm$0.43 & 28.55$\pm$16.74 & 91.85$\pm$0.51 & 91.62$\pm$0.42 & 5.76 & 220.33 & 40.09 \\
    \midrule
       UGradSL & 11.33$\pm$4.17 & 23.08$\pm$11.53 & 87.86$\pm$3.85 & 87.68$\pm$3.81 & 7.13 & 209.95 & 2.17 \\
        UGradSL+ & 15.63$\pm$8.01 & 26.95$\pm$25.80 & 89.17$\pm$5.86 & 88.29$\pm$5.75 & 7.55 & 220.04 & 51.41\\
    \midrule
       UGradSL (R) & 6.29$\pm$1.41 & 5.73$\pm$3.50 & 93.44$\pm$0.14 & 92.80$\pm$0.27 & \textbf{1.11} & 198.26 & 2.17 \\
        UGradSL+ (R) & 6.12$\pm$0.31 & 5.54$\pm$0.34 & 92.79$\pm$0.01 & 92.49$\pm$0.04 & 1.78 & 196.94 & 51.41 \\
    \bottomrule
    \end{tabular}
    }
\end{table*}

\subsection{Experiment Results}
\textbf{Class-Wise Forgetting} We select the class randomly and run class-wise forgetting on four datasets. We report the results of CIFAR-10 and ImageNet in Table
\ref{tab:class_wise}. The results of CIFAR-100 and SVHN are given in Appendix. 
As we can see, UGradSL and UGradSL+ can boost the performance of GA and FT, respectively without an increment in RTE or drop in TA and RA, leading to comprehensive satisfaction in the main metrics, even in the randomness on $D_f$, showing the robustness and flexibility of our methods in MU regardless of the size of the dataset and the data modality. 
Moreover, in terms of average gap, the proposed method shows its similarity to the retrained model. As for the sum metrics, the proposed method is almost the highest, showing the capability to balance the forgetting metrics (UA, $\mbox{MIA}_{\mbox{Score}}$) and the remaining metrics (TA, RA).
\textbf{Random Forgetting} We select data randomly from every class as ${D}_f$, making sure all the classes are selected and the size of ${D}_f$ is 10\% of the ${D}_{tr}$.
We report the results of CIFAR-100 and 20 NewsGroup in Table \ref{tab:random}. 
Compared with class-wise forgetting, it is harder to improve the MU performance without a significant drop in the remaining accuracy in random forgetting across all dataset. 
When the optimization goal is complete unlearning, our methods are much better in UA and $\mbox{MIA}_{\mbox{Score}}$ than the baseline methods, even retrain. 
Compared with FT and GA, UGradSL+ can improve the UA or $\mbox{MIA}_{\mbox{Score}}$ by more than 50\% with a drop in RA or TA by 15\% at most.
When the optimization goal is to make the approximate MU close to the retrained model, our methods still outperform other baseline methods.
The rest of the experiments are given in Appendix. 
\textbf{Group Forgetting} Although group forgetting can be seen as part of random forgetting, we want to highlight its use case here due to its practical impacts on  \textit{e.g.}, facial attributes classification. 
The identities can be regarded as the subgroup in the attributes. 


\textbf{CIFAR-10 and CIFAR-100} share the same image dataset while CIFAR-100 is labeled with 100 sub-classes and 20 super-classes \citep{krizhevsky2009learning}. 
We train a model to classify 20 super-classes using CIFAR-100 training set. 
The setting of the \textit{group forgetting within one super-class} is to remove one sub-class from one super-class in CIFAR-100 datasets. 
For example, there are five fine-grained fishes in the \textit{Fish} super-class and we want to remove one fine-grained fish from the model.
Different from class-wise forgetting, we do not modify the testing set.
We report the group forgetting in Table~\ref{tab:cifar20}.


\textbf{CelebA} We select CelebA dataset as another real-world case and show the results in Table~\ref{tab:celebA}. We train a binary classification model to classify whether the person is smile or not. There are 8192 identities in the training set and we select 1\% of the identities (82 identities) as $D_f$. Both smiling and non-smiling images are in $D_f$. 
This experiment has significant practical meaning, since the bio-metric, such as identity and fingerprint, needs more privacy protection \citep{minaee2023biometrics}. 
Compared with baseline methods, our method can forget the identity information better without forgetting too much remaining information in the dataset.
This paradigm provides a practical usage of MU and our methods provide a faster and more reliable way to improve the MU performance.

\subsection{Discussion}
\label{sec:discssion}

\textbf{The Discussion of the Performance TradeOff}
As shown in Table \ref{tab:class_wise}, while we achieve strong MU performance there is no apparent drop of the remaining performance (RA, TA) for the class-wise forgetting compared with other baselines.
For random forgetting and group forgetting, the boost of the forgetting performance (UA, $\mbox{MIA}_{\mbox{Score}}$) inevitably leads to drops in the remaining performance as shown in Table \ref{tab:random}, \ref{tab:cifar20}, \ref{tab:celebA}. This is a commonly observed trade-off in the MU literature \citep{jia2023model, warnecke2021machine, graves2021amnesiac}. 
Yet we would like to note that at the mild cost of the drops, we observe significant improvement in MU performance when the unlearning metrics is complete unlearning. For example, in Table \ref{tab:random}, the remaining performance drop in CIFAR-100 are 7.73 (TA) and 9.42 (RA), respectively compared with retrain. However, our unlearning performance boosts are 48.18 (UA) and 52.82 ($\mbox{MIA}_{\mbox{Score}}$), respectively compared with the best baseline methods. 
Even when the metrics is based on the retrained model, the proposed method still can make good balance among all the metrics. 

\textbf{Influence Function in Deep Learning} Influence function is proposed for the convex function. As given in Section \ref{sec:ga_help}, we apply the influence function to the converged model, which can be regarded as a local convex model.
A plot of loss landscape of the retrained model $\btheta_{r}$ on CIFAR-10 dataset is attached to Figure \ref{fig:loss_landscape} in the Appendix. 


\textbf{MIA as a Proxy for "Forgetfulness".} 
Given a model $\theta_\star$, we can evaluate the degree of its generalization by running a membership inference attack on the model.
In the context of the current work, generalization is equivalent to the degree of "forgetfulness" that the forgetting algorithm achieves.
Given the distribution of model response observations $\mathcal{A}_f = \mathcal{A}(\theta_\star, \mathcal{D}_f)$ and $\mathcal{A}_{te} = \mathcal{A}(\theta_\star, \mathcal{D}_{te})$, where $\mathcal{A}$ is an adversary and $A$ is the observation visible to $\mathcal{A}$, one can get the degree of generalization by analyzing the observations.
There are several different ways of analyzing the observations (see Table \ref{tab:mia_scores_ref} for some), but in the context of MU, the most straightforward way is to get the accuracy of $\mathcal{A}$ on the seen and unseen samples ($\mathcal{D}_{te}$ and $\mathcal{D}_{f}$ respectively.
This could be done by computing the $(TP+TN)/(|D_f| + |D_{te}|)$, where the true positive predictions $TP$ correspond to "seen" samples, and true negative predictions $TN$ are "unseen" samples.
We conducted the experiments on CIFAR-10 both for class-wise and random forgetting. The results are given in Table \ref{tab:new_MIA}.
We assume that the distribution of $D_{tr}$ and $D_{te}$ should be the same.
For class-wise forgetting, the additional MIA is almost 1 because $D_f$ is a separate single class and the distribution of $D_f$ and $D_{te}$ without the corresponding class are totally different. 
For random forgetting, the additional MIA is almost 0.5 because $D_f$ is randomly selected from $D_{tr}$ and the distribution of $D_f$ and $D_{te}$ should the same. 
The plots of loss distribution for random and class-wise forgetting are given in Figure \ref{fig:loss_dist} in the Appendix.

\begin{table*}[htb]
    \centering
    \caption{Results of class-wise forgetting and random forgetting on CIFAR-10 with additional MIA.
    }
    \label{tab:new_MIA}
    \scalebox{0.75}{
    \begin{tabular}{c|cccc|c|cc|c}
    \toprule
    Class-wise & UA & $\mbox{MIA}_{\mbox{Score}}$ & RA & TA & Additional MIA & Avg. Gap ($\downarrow$) & Sum. ($\uparrow$) & RTE ($\downarrow$, min)\\
    \midrule
    Retrain & 100.00$\pm$0.00 & 100.00$\pm$0.00 & 98.19$\pm$3.14 & 94.50$\pm$0.34 & 99.23$\pm$0.08 & - & 392.69 & 24.62\\
    \midrule
    FT & 22.71$\pm$5.31 & 79.21$\pm$8.60 & 99.82$\pm$0.09 & 94.13$\pm$0.14 & 99.09$\pm$0.07 & 25.02 & 295.87 & 2.02 \\
    GA & 25.19$\pm$11.38 & 73.48$\pm$9.68 & 96.84$\pm$0.58 & 73.10$\pm$1.62 & 99.43$\pm$0.09 & 31.02 & 268.61 & \textbf{0.08} \\
    IU & 83.92$\pm$1.16 & 92.59$\pm$1.41 & 98.77$\pm$0.12 & 92.64$\pm$0.23 & 99.71$\pm$0.07 & 6.48 & 367.92 & 1.18\\
    BE & 64.93$\pm$0.01 & 98.19$\pm$0.00 & 99.47$\pm$0.00 & 94.00$\pm$0.11 & 99.60$\pm$0.02 & 9.67 & 356.59 & 0.20 \\
    BS & 93.69$\pm$4.32 & 99.82$\pm$0.04 & 97.69$\pm$1.29 & 92.89$\pm$1.26 & 99.56$\pm$0.10 & 2.15 & 384.08 & 0.29 \\
    $\ell_1$-sparse & 100.00$\pm$0.00 & 100.00$\pm$0.00 & 97.86$\pm$1.29 & 96.11$\pm$1.26 & 99.02$\pm$0.15 & 0.48 & 393.97 & 1.00 \\
    SCRUB & 100.00$\pm$0.00 & 100.00$\pm$0.00 & 99.93$\pm$0.01 & 95.22$\pm$0.07 & 100.00$\pm$0.00 & 0.62 & 395.15 & 1.09 \\
    Random Label & 99.99$\pm$0.01 & 100.00$\pm$0.00 & 100.00$\pm$0.00 & 95.50$\pm$0.11 & 99.08$\pm$0.07 & 0.71 & \textbf{395.49} & 1.04 \\
    \midrule
    UGradSL & 94.99$\pm$4.35 & 97.95$\pm$1.78 & 95.47$\pm$4.08 & 86.78$\pm$5.68 & 99.94$\pm$0.01 & 4.38 & 375.19 & 0.22 \\
    UGradSL+ & 100.00$\pm$0.00 & 100.00$\pm$0.00 & 99.26$\pm$0.01 & 94.29$\pm$0.07 & 100.00$\pm$0.00 & \textbf{0.32} & 392.55 & 3.07 \\
    \midrule
    Random &UA & One-side MIA & RA & TA & Additional MIA & Avg. Gap ($\downarrow$) & Sum. ($\uparrow$) & RTE ($\downarrow$, min) \\
    \midrule
    Retrain & 8.07$\pm$0.47 & 17.41$\pm$0.69 & 100.00$\pm$0.01 & 91.61$\pm$0.24 & 50.69$\pm$0.73 & - & 217.09 & 24.66 \\
    \midrule
    FT & 1.10$\pm$0.19 & 4.06$\pm$0.41 & 99.83$\pm$0.03 & 93.70$\pm$0.10 & 54.05$\pm$0.31 & 5.65 & 198.69 & 1.58 \\
    GA & 0.56$\pm$0.01 & 1.19$\pm$0.05 & 99.48$\pm$0.02 & 94.55$\pm$0.05 & 55.04$\pm$0.66 & 6.80 & 195.78 & \textbf{0.31} \\
    IU & 17.51$\pm$2.19 & 21.39$\pm$1.70 & 83.28$\pm$2.44 & 78.13$\pm$2.85 & 53.98$\pm$0.55 & 10.91 & 200.31 & 1.18\\
    BE & 0.00$\pm$0.00 & 0.26$\pm$0.02 & 100.00$\pm$0.00 & 95.35$\pm$0.18 & 55.41$\pm$0.49 & 7.24 & 195.61 & 3.17\\
    BS & 0.48$\pm$0.07 & 1.16$\pm$0.04 & 99.47$\pm$0.01 & 94.58$\pm$0.03 & 55.88$\pm$0.72 & 6.84 & 195.69 & 1.41 \\
    $\ell_1$-sparse & 1.21$\pm$0.38 & 4.33$\pm$0.52 & 97.39$\pm$0.31 & 95.49$\pm$0.18 & 55.01$\pm$0.49 & 6.61 & 198.42 & 1.82 \\
    SCRUB & 0.70$\pm$0.59 & 3.88$\pm$1.25 & 99.59$\pm$0.34 & 94.22$\pm$0.26 & 55.33$\pm$0.59 & 5.98 & 198.42 & 4.05\\
    Random Label & 2.80$\pm$0.37 & 18.59$\pm$3.48 & 99.97$\pm$0.01 & 94.08$\pm$0.12 & 52.17$\pm$0.87 & 2.24 & 198.39 & 1.98\\
    \midrule
    UGradSL & 20.77$\pm$0.75 & 35.45$\pm$2.85 & 79.83$\pm$0.75 & 73.94$\pm$0.75 & 56.27$\pm$0.11 & 17.15 & 209.99 & 0.45\\
    UGradSL+ & 25.13$\pm$0.49 & 37.19$\pm$2.23 & 90.77$\pm$0.20 & 84.78$\pm$0.69 & 56.41$\pm$0.32 & 13.23 & \textbf{237.87} & 3.07\\
    \midrule
    UGradSL (R) & 5.87$\pm$0.51 & 13.33$\pm$0.70 & 98.82$\pm$0.28 & 92.17$\pm$0.23 & 53.54$\pm$0.97 & \textbf{2.01} & 210.19 & 0.45 \\
    UGradSL+ (R) & 6.03$\pm$0.17 & 10.65$\pm$0.13 & 99.79$\pm$0.03 & 93.64$\pm$0.16 & 52.29$\pm$0.85 &  2.76 & 210.11 & 3.07 \\
    \bottomrule
    \end{tabular}
    }
\end{table*}

The ablation study of the smooth rate, the Streisand effect \cite{jansen2015streisand},  gradient analysis as well as the difference between UGrad and UGradSL+ are given in Section \ref{appendix:ablation}, \ref{appendix:streisand_effect}, \ref{appendix:grad_analysis} and \ref{appendix:difference}, respectively. 
As mentioned in Section \ref{sec:exp_res}, there are two unlearning objective for random forgetting. We report the UGradSL and UGradSL+ with the optimization objective based on retraining which are tagged with $(R)$. For a fair comparison, we also try our best to make the baseline methods as close as to retraining. The results of CIFAR-10 and CIFAR-100 for random forgetting are given in Table \ref{tab:faircom} in the Appendix.




\section{Conclusions}

We have proposed UGradSL, a plug-and-play, efficient, gradient-based MU method using smoothed labels. Theoretical proofs and extensive numerical experiments have demonstrated the effectiveness of the proposed method in different MU evaluation metrics. 
Our work has limitations. For example, we desire an efficient way to find the exact MU state in experiments and further explore the applications of MU to promote privacy and fairness. Our method can be further validated and tested in other tasks, such as unlearning recommendation systems, etc. 

\bibliographystyle{unsrt}  
\bibliography{references}  

\clearpage
\newpage

\newpage
\section*{Appendix}
\paragraph{Roadmap} The appendix is composed as follows. Section \ref{app:notation} presents all the notations and their meaning we use in this paper.
Section \ref{sec:framework_app} shows the pipeline of our methods.
Section \ref{sec:app_proof} gives the proof of our theoretical analysis. 
Section \ref{app:exp} shows the additional experiment results with more details that are not given in the main paper due to the page limit.

\section{Notation Table}
\label{app:notation}


The notations we use in the paper is summaried in the Table \ref{tab:notations}.

\begin{table}[h!]
    \centering
    \caption{Notation used in this paper}
    \scalebox{1}{
    \begin{tabular}{l|l}
    \toprule
        Notations & Description \\
    \midrule
        $K$ & The number of class in the dataset \\
        $\mathcal{D}, \mathcal{X}, \mathcal{Y}$ & The general dataset distribution, the feature space and the label space \\
        $D$ & The dataset $D \in \mathcal{D}$ \\
        $D_{tr}, D_{r}, D_{f}$ & The training set, remaining set and forgetting set \\
        $\Theta_{\mathcal{M}}$ & The distribution of models learned using mechanism $\mathcal{M}$ \\
        $\btheta$ & The model weight \\
        $\btheta^{*}$ & The optimal model weight \\
        $\btheta^*_{f, \text{LS}}$ & The optimal model weight trained with $D_f$ whose label is smoothed \\
        $||\btheta||$ & The 2-norm of the model weight \\
        $n$ & The size of dataset \\
        $\varepsilon$ & The up-weighted weight of datapoint $z$ in influence function \\
        $\mathcal{I}(z)$ & Influence function of data point $z$ \\
        $h_{\btheta}$ & A function $h$ parameterized by $\btheta$ \\
        $\ell(h_{\btheta}, z_i)$ & Loss of $h_{\btheta}(x_i)$ and $y_i$ \\
        $R_{tr}(\btheta)$ & The empirical risk of training set when the model weight is $\btheta$\\
        $R_{f}(\btheta)$ & The empirical risk of forgetting set when the model weight is $\btheta$\\
        $R_{r}(\btheta)$ & The empirical risk of remaining set when the model weight is $\btheta$\\
        $H_{\btheta}$ & The Hessian matrix w.r.t. $\btheta$ \\
        $\nabla_{\btheta}$ & The gradient w.r.t. $\btheta$ \\
        $B$ & Data batch \\
        $B^{\text{LS}, \alpha}$ & The smoothed batch using $\alpha$ \\
        $z_i = (x_i, y_i)$ & A data point $z_i$ whose feature is $x_i$ and label is $y_i$ \\
        $\boldsymbol{y}_i$ & The one-hot encoded vector form of $y_i$ \\
        $\boldsymbol{y}_i^{\text{GLS}, \alpha}$ & The smoothed one-hot encoded vector form of $y_i$ where the smooth rate is $\alpha$ \\
        $\alpha$ & Smooth rate in general label smoothing \\
    \bottomrule
    \end{tabular}
    }
    \label{tab:notations}
\end{table}

\section{Related Work}
\paragraph{Differential Privacy} A multitude of \textbf{privacy risk assessment} tools have been proposed to gauge the degree of leakage associated with the training data. Specifically targeted at the training data, model attacks are often used as a proxy metric for privacy leakage in pretrained models. For example, model inversion attacks are designed to extract aggregate information about specific sub-classes rather than individual samples \cite{Fredrikson_2015}. Data extraction attacks aim to reverse engineer individual samples used during training \cite{Carlini_2020}, while property inference attacks focus on inferring properties of the training data \cite{Ganju_2018}.

More relevant to the current work are \textbf{Membership Inference Attacks} (MIA), which predict whether a particular sample was used to train the model. First introduced by Homer et al. \cite{Homer_2008}, membership attack algorithms were later formalized in the context of DP, enabling privacy attacks and defenses for machine learning models \cite{Rahman_2018}. Shokri et al. \cite{Shokri_2017} introduced MIA based on the assumption of adversarial queries to the target model. By training a reference attack model (shadow model) based on the model inference response, this type of MIA has proven to be powerful in scenarios such as white-box \cite{Leino_2019,Nasr_2019,Sablayrolles_2019}, black-box \cite{Chen_2020,Hisamoto_2019,Song_2020}, and label-only \cite{Choquette-Choo_2020,Li_2021} access. However, most MIA mechanisms often require training a large number of shadow models with diverse subsets of queries, making them prohibitively expensive. As a result, some recent works have focused on developing cheaper MIA mechanisms \cite{Steinke_2023}.

Adopted from \cite{Zarifzadeh_2023_2}, table \ref{tab:mia_scores_ref} shows the summary of MIA scores that could be used as a proxy for evaluating the effectiveness of "forgetting". 
Note that in the context of \cite{Zarifzadeh_2023_2} the "MIA Score" is used to perform an adversarial attack such that $\mbox{MIA}(x, \theta)=\mathbb{I}_{\mbox{Score}(x,\theta)\ge\beta}$ (with $\mathbb{I}$ being the characteristic function). In the context of the current work, we consider the said scores for the purpose of evaluating the "forgetness".

\begin{table}[htb]
    \centering
    \caption{Computation of the MIA-based membership exposure score.}
    \label{tab:mia_scores_ref}
    \scalebox{1}{
    \begin{tabular}{lccccc}
        \toprule
        Method & RMIA \cite{Zarifzadeh_2023_2} & LiRA \cite{Carlini_2022} & Attack-R \cite{Ye_2022} & Attack-P \cite{Ye_2022} & Global \cite{Yeom_2018} \\
        \midrule
        MIA Score & $\mathbb P_z\left(\frac{\mathbb P(\theta|x)}{\mathbb P(\theta|z)}\right)$ & 
            $\frac{\mathbb P(\theta|x)}{\mathbb P(\theta|\bar{x})}$ & 
            $\mathbb P_{\theta'}\left(\frac{\mathbb P(x|\theta)}{\mathbb P(x|\theta')}\right) \ge 1$ & 
            $\mathbb P_{z}\left(\frac{\mathbb P(x|\theta)}{\mathbb P(z|\theta)}\right) \ge 1$ & 
            $\mathbb P\left(x|\theta\right)$ \\
        \bottomrule
    \end{tabular}
    }
\end{table}

\textbf{Basics of Influence Function}
Given a dataset $D = \{z_i:(x_i, y_i)\}_{i=1}^n$ and a function $h$ parameterized by $\btheta$ which maps from the input feature space $\mathcal{X}$ to the output space $\mathcal{Y}$. 
The standard empirical risk minimization writes as: 
\begin{equation}
\label{equ:min_erm}
    \btheta^*=\arg \min _{\btheta} \frac{1}{n} \sum_{z \in D} \ell\left(h_{\btheta}, z\right).
\end{equation}
To find the impact of a training point $\hat{z}$, we up-weight its weight by an infinitesimal amount $\varepsilon$\footnote{To distinguish from the $\epsilon$ in differential privacy, we use $\varepsilon$ here.}. The new model parameter $\btheta_{\{\hat{z}\}}^{\varepsilon}$ can be obtained from 
\begin{equation}
    \btheta_{\{z\}}^{\epsilon_I}=\arg \min _{\btheta} \frac{1}{n} \sum_{z \in D} \ell\left(h_{\btheta}, z\right)+\varepsilon \cdot \ell\left(h_{\btheta}, \hat{z}\right)
\end{equation}

When $\varepsilon = -\frac{1}{n}$, it is indicating removing $\hat{z}$. According to \cite{koh2017understanding}, $\btheta_{\{\hat{z}\}}^{\varepsilon}$ can be approximated by using the first-order Taylor series expansion as 
\begin{equation}
    \btheta_{\{\hat{z}\}}^{\varepsilon} \approx \btheta^*-\varepsilon \cdot H_{\btheta^*}^{-1}\cdot \nabla_{\btheta} \ell\left(h_{\btheta^*}, \hat{z}\right),
\end{equation}

where $H_{\btheta^*}$ is the Hessian with respect to (w.r.t.) $\btheta^{*}$. The change of $\btheta$ due to changing the weight can be given using the influence function $\mathcal{I}(\hat{z})$ as 
\[
    \Delta \btheta = \btheta^{\varepsilon}_{\{\hat{z}\}} - \btheta^{*} = \mathcal{I}(\hat{z})=\left.\frac{d \btheta_{\{\hat{z}\}}^{\varepsilon}}{d \varepsilon}\right|_{\varepsilon=0}=-H_{\btheta^*}^{-1} \cdot \nabla_{\btheta} \ell\left(h_{\btheta^*}, \hat{z}\right).
\]

\section{The Framework of Our Method}
\label{sec:framework_app}
\begin{figure}[ht]
    \centering
    \includegraphics[width=0.9\textwidth]{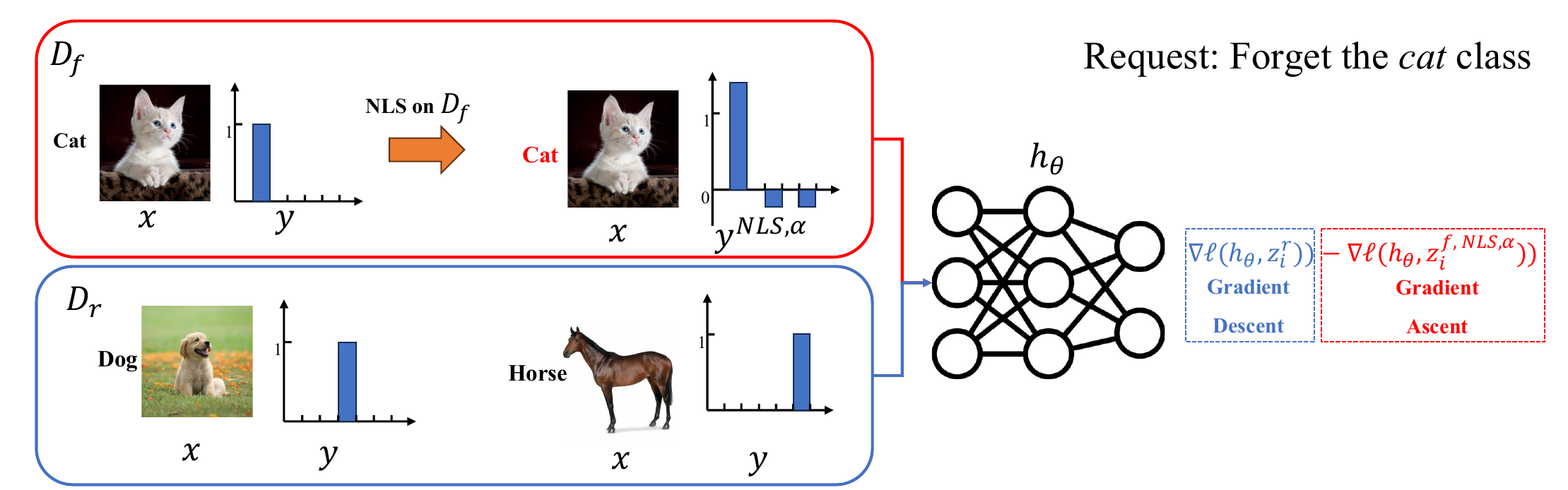}
    \caption{The framework of UGradLS. When there is a unlearning request, we can split the $D_{tr}$ into $D_f$ and $D_r$. We first apply NLS on $z_i^f=\{x, y\} \in D_f$ to get $z_i^{\text{NLS}, \alpha}=\{x, y^{\text{NLS}, \alpha}\}$. In back-propagation process, we apply gradient descent on the data $z_{i}^r \in D_r$ and gradient ascent on the data smoothed $D_f$, which is the mix-gradient way. }
    \label{fig:teaser}
\end{figure}

Our framework is shown in given Figure \ref{fig:teaser}. We only apply NLS on the forgetting dataset $D_f$. In back-propagation process, we apply gradient descent on the data $z_{i}^r \in D_r$ and gradient ascent on the data smoothed $D_f$, which is the mix-gradient way.

\begin{figure}[ht]
    \centering
    \includegraphics[width=\textwidth]{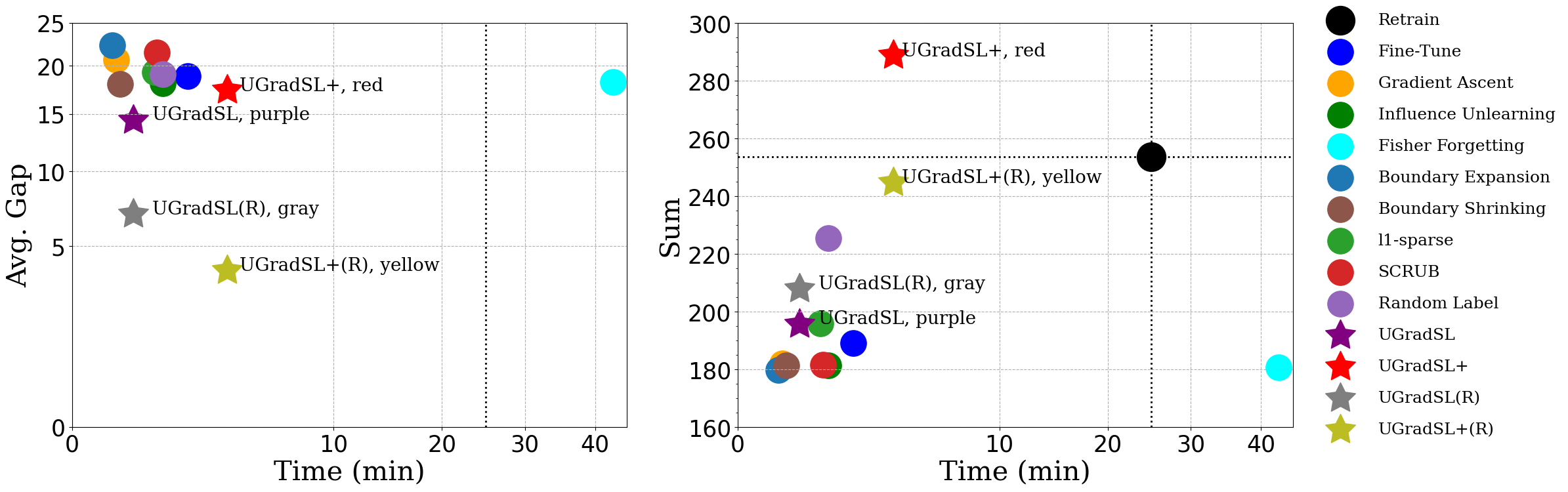}
    \caption{Summary of the proposed method and baselines (CIFAR 100, random forgetting across all classes) in terms of the two different MU metrics and speed. The proposed method with a following \textit{(R)} means the method is optimized according to the retrained model while the method without a \textit{(R)} means the method is optimized according to the \textit{completely unlearning} \citep{shah2023unlearning}. 
    For the plot of \textbf{Average Gap}, the \textbf{lower left} indicates better performance. Average Gap is to evaluate how close the approximate MU method is to the retrained model. The smaller the average gap is, the better MU performance is. The retrained model has no average gap as reference in this plot. The \textbf{bold} vertical dash shows the speed. When we target to make the approximate MU method close to the retrained model, UGradSL ($\color{violet}\filledstar$), UGradSL(R, $\color{gray}\filledstar$) and UGradSL+(R, $\color{yellow}\filledstar$) outperform the other methods.
    For the plot of \textbf{Sum}, the \textbf{upper left} indicates better performance. Sum is to evaluate the comprehensive performance of the approximate MU model. The higher the sum is, the better performance of the MU method is. The black dot represents the retrained model. When we target to improve the comprehensive performance of the approximate MU, UGradSL+($\color{red}\filledstar$) and UGradSL+(R, $\color{yellow}\filledstar$) outperform the other methods without too much delay. 
    Moreover, in both plots, our proposed methods are always the top methods.
    Our methods show the robustness and generalization capability in both MU metrics.}
    \label{fig:my_label}
\end{figure}

\section{Proofs}
\label{sec:app_proof}

\subsection{Proof for Theorem~\ref{the:GA_helps}}
\label{sec:proof_GA_helps}
\begin{proof}
For $p(x)$, the Taylor expansion at $x=a$ is 
\begin{equation}
    p(x) = p(a) + \frac{p^{\prime}(a)}{1}(x-a) + o
\end{equation}

Here $p(\btheta) = \nabla R_{tr}({\btheta}) + \varepsilon \sum_{D_{f}} \nabla \ell(h_{\btheta}, z^{f}_{i})$ so we have 
\begin{equation}
\begin{split}
    p(\btheta) &= \nabla R_{tr}(a) + \varepsilon \sum_{z^{f} \in D_{f}} \nabla \ell(h_a, z^{f}) + (\nabla^2 R_{tr}({a}) + \varepsilon \sum_{z^{f} \in D_{f}} \nabla^2 \ell(h_a, z^{f})(\btheta - a) + o \\
\end{split}
\end{equation}

For Equation~\ref{equ:unlearning_opt}, we expand $f(\btheta_f)$ at $\btheta = \btheta^{*}_{tr}$ as 
\begin{equation}
    \begin{split}
        p(\btheta^{*}_{f}) &= \nabla R_{tr}(\btheta^{*}_{tr}) + \varepsilon \sum_{z^{f} \in D_{f}} \nabla \ell(h_{\btheta^{*}_{tr}}, z^{f}) + \left[\nabla^2 R_{tr}({\btheta^{*}_{tr}}) + \varepsilon \sum_{z^{f} \in D_{f}} \nabla^2 \ell(h_{\btheta^{*}_{tr}}, z^{f})\right] (\btheta^{*}_{f} - \btheta^{*}_{tr}) + o = 0 \\
        & \nabla R_{tr}(\btheta^{*}_{tr}) + \varepsilon \sum_{z^{f} \in D_{f}} \nabla \ell(h_{\btheta^{*}_{tr}}, z^{f}) + \left[\nabla^2 R_{tr}({\btheta^{*}_{tr}}) + \varepsilon \sum_{z^{f} \in D_{f}} \nabla^2 \ell(h_{\btheta^{*}_{tr}}, z^{f})\right](\btheta^{*}_{f} - \btheta^{*}_{tr})  \approx 0 \\
        & -\left[\nabla^2 R_{tr}({\btheta^{*}_{tr}}) + \varepsilon \sum_{z^{f} \in D_{f}} \nabla^2 \ell(h_{\btheta^{*}_{tr}}, z^{f})\right]^{-1}\left[\nabla R_{tr}(\btheta^{*}_{tr}) + \varepsilon \sum_{z^{f} \in D_{f}}\nabla \ell(h_{\btheta^{*}_{tr}}, z^{f})\right] = (\btheta^{*}_{f} - \btheta^{*}_{tr}) \\
    \end{split}
\end{equation}

We have $\nabla R_{tr}(\btheta_{tr}^{*}) = 0$ and 
\begin{equation}
    \begin{split}
        \btheta^{*}_{f} - \btheta^{*}_{tr} &= -\left[\nabla^2 R_{tr}({\btheta^{*}_{tr}}) + \varepsilon \sum_{z^{f} \in D_{f}} \nabla^2 \ell(h_{\btheta^{*}_{tr}}, z^{f})\right]^{-1} \left( \varepsilon \sum_{z^{f} \in D_{f}} \nabla \ell(h_{\btheta^{*}_{tr}}, z^{f}) \right) \\
        &= -\left[\sum_{z^{tr} \in D_{tr}} \nabla^{2} \ell(h_{\btheta_{tr}^{*}}, z^{tr}) + \varepsilon \sum_{z^{f} \in D_{f}}\nabla^2 \ell(h_{\btheta^{*}_{tr}}, z^{f})\right]^{-1} \left( \varepsilon \sum_{z^{f} \in D_{f}} \nabla \ell(h_{\btheta^{*}_{tr}}, z^{f}) \right)
    \end{split}
\end{equation}

We expand $q(\btheta^{*}_{tr})=R_{tr}(\btheta^{*}_{tr})=\sum_{z^{tr} \in D_{tr}} \ell(h_{\btheta}, z^{tr})$ at $\btheta=\btheta^{*}_{r}$ as 
\begin{equation}
    \begin{split}
        q(\btheta_{tr}^{*}) &= \sum_{z^{tr} \in D_{tr}} \nabla \ell(h_{\btheta_{r}^{*}}, z^{tr}) + \sum_{z^{tr} \in D_{tr}} \nabla^2 \ell(h_{\btheta_{r}^{*}}, z^{tr})(\btheta^{*}_{tr} - \btheta_{r}^{*}) \approx 0 \\
        \btheta^{*}_{r} - \btheta^{*}_{tr} & = \left(\sum_{z^{tr} \in D_{tr}} \nabla^2 \ell(h_{\btheta_{r}^{*}},  z^{tr})\right)^{-1}\sum_{z^{tr} \in D_{tr}} \nabla \ell(h_{\btheta_{r}^{*}}, z^{tr}) \\
    \end{split}
\end{equation}

Because of gradient ascent, $\epsilon = -1$ and we have
\begin{equation}
\label{equ:r_k_diff}
    \begin{split}
        \btheta_r^{*} - \btheta_{f}^{*} &= \btheta^*_r - \btheta^*_{tr} - (\btheta^*_{tr}-\btheta^*_{f}) = \left(\sum_{z^{tr} \in D_{tr}} \nabla^2 \ell(h_{\btheta_{r}^{*}}, z^{tr})\right)^{-1}\sum_{z^{tr} \in D_{tr}} \nabla \ell(h_{\btheta_{r}^{*}}, z^{tr}) \\
        &-\left[\sum_{z^{tr} \in D_{tr}} \nabla^{2} \ell(h_{\btheta_{tr}^{*}}, z^{tr}) - \sum_{z^{f} \in D_{f}}\nabla^2 \ell(h_{\btheta^{*}_{tr}}, z^{f})\right]^{-1} \sum_{z^{f} \in D_{f}} \nabla \ell(h_{\btheta^{*}_{tr}}, z^{f}) \\
        &= \underbrace{\left(\sum_{z^{tr} \in D_{tr}} \nabla^2 \ell(\theta_{r}^{*}, z^{tr})\right)^{-1}\sum_{z^{tr} \in D_{tr}} \nabla \ell(\theta_{r}^{*}, z^{tr})}_{\Delta \btheta_{r}} - \underbrace{\left(\sum_{z^{r} \in D_r}\nabla^2 \ell(\theta_{tr}^{*}, z^{r})\right)^{-1}\sum_{z^{f} \in D_{f}}\nabla \ell(\theta^{*}_{tr}, z^{f})}_{\Delta \btheta_{f}} \\
    \end{split}
\end{equation}

Thus, $||\btheta_r^{*} - \btheta_f^{*}|| = 0$ if and only if
\begin{equation}
    \sum_{z^{f} \in D_{f}} \nabla_{\btheta} \ell(h_{\btheta_{r}^{*}}, z^{f}) =  - \left(\sum_{z^{tr} \in D_{tr}} \nabla_{\btheta}^2 \ell(h_{\btheta_{r}^{*}}, z^{tr})\right) \left(\sum_{z^{r} \in D_r}\nabla^2_{\btheta} \ell(h_{\btheta_{tr}^{*}}, z^{r})\right)^{-1} \sum_{z^{f} \in D_{f}}\nabla_{\btheta} \ell(h_{\btheta^{*}_{tr}}, z^{f})
\end{equation}
\end{proof}

\subsection{Proof for Theorem~\ref{the:negative_is_pos}}
\label{sec:negative_is_pos}
\begin{proof}

Recall the loss calculation in label smoothing and we have
\begin{equation}
\label{equ:gls_app}
    \ell(h_{\btheta}, z^{\text{GLS}, \alpha}) = (1+\frac{1-K}{K}\alpha)\ell(h_{\btheta}, (x, y)) + \frac{\alpha}{K}\sum_{y^{\prime} \in \mathcal{Y} \backslash y} \ell(h_{\btheta}, (x, y^{\prime}))),
\end{equation}
where we use notations $\ell(h_{\btheta}, (x, y)):=\ell(h_{\btheta}, z)$ to specify the loss of an example $z=\{x, y\}$ existing in the dataset and $\ell(h_{\btheta}, (x, y^{\prime}))$ to denote the loss of an example when its label is replaced with $y^{\prime}$. $\nabla_{\btheta} \ell(h_{\btheta}, (x, y))$ is the gradient of the target label and $\sum_{y^{\prime} \in \mathcal{Y}\backslash y} \nabla_{\btheta} \ell(h_{\btheta}, (x, y^{\prime}))$ is the sum of the gradient of non-target labels. 

With label smoothing in Equation~\ref{equ:gls_app}, Equation~\ref{equ:r_k_diff} becomes


\begin{equation}
\label{equ:param_distance_NLS_dev}
\begin{split}
    & \btheta_{r}^{*} - \btheta_{f,\text{LS}}^{*} \\
    & \approx \Delta \btheta_{r}  + (1+\frac{1-K}{K}\alpha) \cdot (-\Delta \btheta_{f})  \\
    & + \frac{1-K}{K}\alpha \cdot \underbrace{\frac{1}{K-1}\left(\sum_{z^{r} \in D_r}\nabla^2_{\btheta} \ell(h_{\btheta_{tr}^{*}}, z^{r})\right)^{-1}\sum_{z^f \in D_{f}}\nabla_{\btheta}  \sum_{y^{\prime} \in \mathcal{Y} \backslash y^f} \ell(h_{\btheta_{tr}^{*}}, (x^f, y^{\prime}))}_{\Delta \btheta_{n}} \\
    &= {\Delta \btheta_{r}}  + (1+\frac{1-K}{K}\alpha) \cdot (-\Delta \btheta_{f})  + \frac{1-K}{K}\alpha \cdot {\Delta \btheta_{n}} \\
    &= {\Delta \btheta_{r}} -{\Delta \btheta_{f}} + \frac{1-K}{K}\alpha \cdot (\Delta \btheta_{n}-{\Delta \btheta_{f}})  \\
\end{split}
\end{equation}

where 
\begin{equation*}
\begin{split}
    \Delta \btheta_{r} &:=\left(\sum_{z^{tr} \in D_{tr}} \nabla_{\btheta}^2 \ell(h_{\btheta_{r}^{*}}, z^{tr})\right)^{-1}\sum_{z^{tr} \in D_{tr}} \nabla_{\btheta} \ell(h_{\btheta_{r}^{*}}, z^{tr}) \\
    \Delta \btheta_{f} &:= \left(\sum_{z^{r} \in D_r}\nabla^2_{\btheta} \ell(h_{\btheta_{tr}^{*}}, z^{r})\right)^{-1}\sum_{z^{f} \in D_{f}}\nabla_{\btheta} \ell(h_{\btheta^{*}_{tr}}, z^{f})
\end{split}
\end{equation*}


as given in Equation~\ref{equ:r_k_diff}. So we have

\begin{equation}
\label{equ:param_distance_NLS}
\begin{split}
    & \btheta_{r}^{*} - \btheta_{f,LS}^{*} \approx  {\Delta \btheta_{r}} -{\Delta \btheta_{f}} + \frac{1-K}{K}\alpha \cdot (\Delta \btheta_{n}-{\Delta \btheta_{f}})  \\
\end{split}
\end{equation}

where 
\begin{equation*}
    \Delta \btheta_{n} := \frac{1}{K-1}\left(\sum_{z^{r} \in D_r}\nabla^2_{\btheta} \ell(h_{\btheta_{tr}^{*}}, z^{r})\right)^{-1}\sum_{z^f \in D_{f}}\nabla_{\btheta}  \sum_{y^{\prime} \in \mathcal{Y} \backslash y^f} \ell(h_{\btheta_{tr}^{*}}, (x^f, y^{\prime}))
\end{equation*}
When we have 
\begin{equation}
   \langle \Delta \btheta_{r} - \Delta \btheta_{f}, \Delta \btheta_n-\Delta \btheta_{f}\rangle \le 0,
\end{equation}

$\alpha < 0$ can help with MU, making
\begin{equation}
    ||\btheta_r^* - \btheta_{f, \text{NLS}}^*|| \leq ||\btheta_r^*-\btheta_f^*||
\end{equation}
\end{proof}


\subsection{Proof for Theorem~\ref{thm:erm}}
\begin{proof}

When the optimization is gradient ascent (GA) with negative label smoothing (NLS), Equation~(\ref{equ:gls}) can be written as
\begin{equation}
    \ell(h_{\btheta}, z^{\text{NLS}, \alpha}) = -\left(1+\frac{1-K}{K}\alpha\right)\cdot \ell(h_{\btheta}, (x,y)) - \frac{\alpha}{K}\sum_{y^{\prime}\in \mathcal{Y} \backslash y} \ell(h_{\btheta}, (x,y^{\prime})), \alpha < 0,
\end{equation}

Recall $R_{tr}(\btheta)=\sum_{z^{tr} \in D_{tr}} \ell(h_{\btheta}, z^{tr})$. Denote by $R^{\text{NLS}}_{f}(\btheta; \alpha)=\sum_{z^{\text{LS},\alpha} \in D_{f}} \ell(h_{\btheta}, z^{\text{NLS},\alpha}), \alpha < 0$ the empirical risk of forgetting data with NLS. After MU with label smoothing on $D_f$ by gradient ascent, the resulting model can be seen as minimizing the risk $\gamma_1 \cdot R_{tr}(\btheta) - \gamma_2 \cdot R^{\text{NLS}}_{f}(\btheta; \alpha)$, which is a weighted combination of the risk from two phases: 1) machine learning on $D_{tr}$ with weight $\gamma_1>0$ and 2) machine unlearning on $D_{f}$ with weight $\gamma_2>0$.
Consider an example $(x,y)$ in the forgetting dataset. The loss of this example is:
\begin{align*}
    &\gamma_1 \ell(h_{\btheta}, (x,y)) - \gamma_2 \ell(h_{\btheta}, z^{\text{GLS}, \alpha})\\
    =&  \left(\gamma_1 - \gamma_2\left(1+\frac{1-K}{K}\alpha\right) \right)\cdot \ell(h_{\btheta}, (x,y)) - \frac{\alpha}{K} \gamma_2\sum_{y^{\prime}\in \mathcal{Y} \backslash y} \ell(h_{\btheta}, (x,y^{\prime})).
\end{align*}


When 
$\left(\gamma_1 - \gamma_2\left(1+\frac{1-K}{K}\alpha\right) \right) > 0$,
the optimal solution by minimizing this loss is 
\begin{align*}
    \mathbb P(\mathcal M(y) = y{^{\texttt{pred}}}) = \left\{ \begin{array}{ll}
\frac{\gamma_1 - \gamma_2\left(1+\frac{1-K}{K}\alpha\right)}{\left(\gamma_1 - \gamma_2\left(1+\frac{1-K}{K}\alpha\right) \right) - \frac{K-1}{K}\alpha \gamma_2}, & \text{if}~y{^{\texttt{pred}}} = y, \\
 \frac{-\frac{\alpha}{K}\cdot \gamma_2}{\left(\gamma_1 - \gamma_2\left(1+\frac{1-K}{K}\alpha\right) \right) - \frac{K-1}{K}\alpha \gamma_2}, & \text{if}~y{^{\texttt{pred}}} \ne y.
\end{array}\right.
\end{align*}

Accordingly, for another label $y'$, we have 
\begin{align*}
    \mathbb P(\mathcal M(y') = y{^{\texttt{pred}}}) = \left\{ \begin{array}{ll}
\frac{\gamma_1 - \gamma_2\left(1+\frac{1-K}{K}\alpha\right)}{\left(\gamma_1 - \gamma_2\left(1+\frac{1-K}{K}\alpha\right) \right) - \frac{K-1}{K}\alpha \gamma_2}, & \text{if}~y{^{\texttt{pred}}} = y', \\
 \frac{-\frac{\alpha}{K}\cdot \gamma_2}{\left(\gamma_1 - \gamma_2\left(1+\frac{1-K}{K}\alpha\right) \right) - \frac{K-1}{K}\alpha \gamma_2}, & \text{if}~y{^{\texttt{pred}}} \ne y'.
\end{array}\right.
\end{align*}
Then the quotient of two probabilities can be upper bounded by:
\[
\log\left(\frac{\mathbb P(\mathcal M(y) = y{^{\texttt{pred}}})}{\mathbb P(\mathcal M(y') = y{^{\texttt{pred}}})} \right) \le \left| \log\left( \frac{\gamma_1 - \gamma_2\left(1+\frac{1-K}{K}\alpha\right)}{-\frac{\alpha}{K}\cdot \gamma_2} \right)\right| = \left| \log\left(  \frac{K}{\alpha} (1-\frac{\gamma_1}{\gamma_2})  + 1 - K \right)\right| = \epsilon.
\]


\end{proof}

\section{The details of Algorithm}
\label{sec:algo_details}
We provide a more detailed explanation of UGradSL and UGradSL+ in Algorithm \ref{alg:cap} here.
For UGradSL+, we first sample a batch $B_{r} = \{z_i^r:(x_i^r, y_i^r)\}_{i=1}^{n_{B_r}}$ from $D_r$ (Line 3-4).
Additionally, we sample a batch $B_{f} = \{z_i^f:(x_i^f, y_i^f)\}_{i=1}^{n_{B_f}}$ from $D_f$ where $n_{B_r} = n_{B_f}$ (Line 5). 
We compute the distance $d(z_i^r, z_i^f) \in [0, 1]$ for each $(z_i^r, z_i^f)$ pair where $z_i^r \in B_r$ and $z_i^f \in B_f$ (Line 6).
For each $z_i^f$, we count the number of $z_i^r$ whose $d(z_i^r, z_i^f)$ < $\beta$, where $\beta$ is the distance threshold. This count is denoted by $c_i^f$ (Line 7).
Then we get the smooth rate by normalizing the count as $\alpha_{i} = c_i^f/|B_f|$, where $\alpha_{i} \in [0,1]$ (Line 8).
GA with NLS is to decrease the model confidence of ${D}_f$. The larger the absolute value of $\alpha_i$, the lower confidence will be given. Our intuition is that a smaller $d(z_i^r, z_i^f)$ means $z_i^r$ is more similar to $D_r$ and the confidence of $z_i^f$ should not be decreased too much. 

\section{Experiments}\label{app:exp}

\subsection{Ablation Study}
\label{appendix:ablation}
Gradient-mixed and NLS are the main contribution to the MU improvement. 
We study the influence of gradient-mixed and NLS on UGradSL and UGradSL+ using random forgetting across all classes in CIFAR-10, respectively. 
Compared with NLS, PLS is a commonly-used method in GLS. We also study the difference between PLS and NLS by replacing NLS with PLS in our methods. The results are shown in Table \ref{tab:abalation}. We can find that gradient-mixed can improve the GA or FT while NLS can improve the methods further. 
\begin{table}[htb!]
    \centering
    \caption{Ablation study of gradient-mixed and NLS using random forgetting across all classes in CIFAR-10. UGradSL can still work without $D_r$, showing the effectiveness of NLS on MU. Gradient-Mixed cannot be removed from UGradSL+ because UGradSL+ without $D_f$ is the same as FT.}
    \scalebox{0.8}{
    \begin{tabular}{c|ccc|cccc|c|c}
    \toprule
    & Gradient-Mixed & NLS & PLS & UA & $\mbox{MIA}_{\mbox{Score}}$ & RA & TA & Sum ($\uparrow$) & RTE ($\downarrow$, min) \\
    \midrule
    GA & & & & 0.56$\pm$0.01 & 1.19$\pm$0.05 & 99.48$\pm$0.02 & 94.55$\pm$0.05 & 195.78 & 0.31 \\
    \midrule
        & & $\checkmark$ & & 25.20$\pm$1.67 & 33.66$\pm$2.11 & 76.41$\pm$1.59 & 70.15$\pm$1.31 & 205.42 & 0.36 \\
        & $\checkmark$ & & & 0.58$\pm$0.00 & 1.18$\pm$0.06 & 99.48$\pm$0.02 & 94.61$\pm$0.05 & 195.85 & 0.46 \\
    UGradSL &    $\checkmark$ & $\checkmark$ & & 20.77$\pm$0.75 & 35.45$\pm$2.85 & 79.83$\pm$0.75 & 73.94$\pm$0.75 & 209.99 & 0.45 \\
    & $\checkmark$ & & $\checkmark$ & 2.02$\pm$0.28 & 18.66$\pm$0.03 & 98.03$\pm$0.37 & 92.15$\pm$0.40 & 210.86 & 0.46\\
    \midrule
    \midrule
    Fine-Tune & & & & 1.10$\pm$0.19 & 4.06$\pm$0.41 & 99.83$\pm$0.03 & 93.70$\pm$0.10 & 198.69 & 1.58  \\
    \midrule
     & $\checkmark$ & & & 14.12$\pm$0.27 & 18.31$\pm$0.07 & 97.31$\pm$0.19 & 90.17$\pm$10.17 & 219.91 & 3.07\\
    UGradSL+ & $\checkmark$ & $\checkmark$ & & 25.13$\pm$0.49 & 37.19$\pm$2.23 & 90.77$\pm$0.20 & 84.78$\pm$0.69 & 237.87 & 3.07 \\
    & $\checkmark$ & & $\checkmark$ & 10.81$\pm$3.76 & 22.29$\pm$0.81 & 93.98$\pm$3.10 & 87.96$\pm$2.68 & 215.04 & 3.01\\
    \bottomrule
    \end{tabular}
    }
    \label{tab:abalation}
\end{table}

\subsection{Baseline Methods}

Retrain is to train the model using $D_r$ from scratch. The hyper-parameters are the same as the original training. 
FT is to fine-tune the original model  $\btheta_o$ trained from $D_{tr}$ using $D_r$. The differences between FT and retrain are the model initialization $\btheta_o$ and much smaller training epochs.
FF is to perturb the $\btheta_o$
by adding the Gaussian noise, which with a zero mean and a covariance corresponds to the 4th root of the Fisher Information Matrix with respect to (w.r.t.) $\btheta_o$ on $D_r$ \citep{golatkar2020eternal}.
IU uses influence function \citep{koh2017understanding} to estimate the change from $\btheta_o$ to $\btheta_u$ when one training sample is removed.
BU unlearns the data by assigning pseudo label and manipulating the decision boundary. 

\subsection{Implementation Details}
\label{sec:implementation_details}
We run all the experiments using PyTorch 1.12 on NVIDIA A5000 GPUs and AMD EPYC 7513 32-Core Processor.
For CIFAR-10, CIFAR-100 and SVHN, the training epochs learning rate are 160 and 0.01, respectively.
For ImageNet, the training epochs are 90.
For 20 Newsgroup, the training epochs are 60.
The batch size is 256 for all the dataset.
Retrain follows the same settings of training. 
For fine-tune (FT), the training epochs and learning rate are 10 and 0.01, respectively.
For gradient ascent (GA), the training epochs and learning rate are 10 and 0.0001, respectively.

\subsection{Class-Wise Forgetting}
We present the performance of class-wise forgetting in CIFAR-100 and SVHN dataset in Table \ref{tab:class_wise_app}. The observation is similar in CIFAR-10 and ImageNet given in Table \ref{tab:class_wise}. UGradSL and UGradSL+ can improve the MU performance with acceptable time increment and performance drop in $D_r$. In addition, UGradSL and UGradSL+ can improve the randomness of the prediction in $D_f$.

\begin{table*}[t]
    \centering
    \caption{The experiment results of class-wise forgetting in CIFAR-100 and SVHN. 
    }
    \label{tab:class_wise_app}
    \scalebox{0.85}{%
    \begin{tabular}{c|cccc|cc|c}
    \toprule
        CIFAR-100 & UA & $\mbox{MIA}_{\mbox{Score}}$ & RA & TA & Avg. Gap ($\downarrow$) & Sum. ($\uparrow$) & RTE ($\downarrow$, min) \\
    \midrule
    Retrain & 100.00$\pm$0.00 & 100.00$\pm$0.00 & 99.96$\pm$0.01 & 71.10$\pm$0.12 & - & 371.05 & 26.95 \\
        \midrule
       FT & 0.67$\pm$0.38 & 27.20$\pm$1.34 & 99.96$\pm$0.01 & 71.46$\pm$0.09 & 43.12 & 199.28 & 1.74 \\
       GA & 99.00$\pm$0.57 & 99.07$\pm$0.50 & 77.83$\pm$2.07 & 53.73$\pm$0.96 & 10.36 & 329.63 & \textbf{0.06} \\
       IU & 2.07$\pm$1.65 & 33.20$\pm$8.83 & 99.96$\pm$0.01 & 71.39$\pm$0.19 & 41.26 & 206.62 & 1.24\\
       BE & 99.07$\pm$0.34 & 99.00$\pm$0.49 & 70.81$\pm$2.69 & 49.85$\pm$1.32 & 13.08 & 318.73 & 0.55\\
       BS & 98.87$\pm$0.57 & 98.73$\pm$0.68 & 71.16$\pm$2.60 & 50.03$\pm$1.36 & 13.06 & 318.80 & 0.77\\
       $\ell_1$-sparse & 98.97$\pm$1.03 & 100.00$\pm$0.00 & 86.99$\pm$0.76 & 79.08$\pm$0.75 & 4.56 & 360.26 & 0.15\\
       SCRUB & 30.07$\pm$49.48 & 66.60$\pm$29.19 & 99.98$\pm$0.01 & 77.97$\pm$0.56 & 26.62 & 274.62 & 1.07\\
       Random Label & 99.80$\pm$0.35 & 100.00$\pm$0.00 & 99.97$\pm$0.62 & 77.31$\pm$0.35 & 0.67 & \textbf{371.18} & 1.10\\
       \midrule
       UGradSL & 66.59$\pm$0.90 & 90.96$\pm$5.05 & 95.45$\pm$1.42 & 70.34$\pm$1.78 & 12.87 & 323.34 & 0.07 \\
       UGradSL+ & 100.00$\pm$0.00 & 100.00$\pm$0.00 & 98.44$\pm$0.62 & 74.12$\pm$0.70 & \textbf{0.57} & 372.56 & 3.37 \\
    \midrule
        SVHN & UA & $\mbox{MIA}_{\mbox{Score}}$ & RA & TA & Avg. Gap ($\downarrow$) & Sum. ($\uparrow$) & RTE ($\downarrow$, min)\\
    \midrule
    Retrain & 100.00$\pm$0.00 &100.00$\pm$0.00 & 100.00$\pm$0.01 & 95.94$\pm$0.11 & - & 395.94 & 37.05\\ 
        \midrule
       FT  & 6.49$\pm$1.49 & 99.98$\pm$0.04 & 100.00$\pm$0.01 & 96.08$\pm$0.01 & 23.42 & 302.55 & 2.42\\
       GA  & 87.49$\pm$1.94 & 99.85$\pm$0.09 & 99.52$\pm$0.03 & 95.27$\pm$0.21 & 3.45 & 388.73 & 0.15\\
       IU  & 93.55$\pm$2.78 & 100.00$\pm$0.00 & 99.54$\pm$0.03 & 95.64$\pm$0.31 & 1.80 & 388.73 & \textbf{0.23}\\
       FF & 72.45$\pm$44.51 & 77.98$\pm$23.99 & 39.36$\pm$41.12 & 37.16$\pm$39.36 & 42.25 & 226.95 & 5.88\\
       BE & 85.56$\pm$3.07 & 99.98$\pm$0.02 & 99.55$\pm$0.01 & 95.53$\pm$0.07 & 3.83 & 380.62 & 3.17\\
       BS & 96.62$\pm$1.14 & 99.95$\pm$0.09 & 99.99$\pm$0.00 & 95.39$\pm$0.18 & 1.00 & 391.95 & 3.91\\
       $\ell_1$-sparse & 99.78$\pm$0.31 & 100.00$\pm$0.00 & 98.63$\pm$0.01 & 97.36$\pm$0.18 & 0.75 & 395.77 & 2.91\\
       SCRUB & 99.99$\pm$0.02 & 100.00$\pm$0.00 & 100.00$\pm$0.00 & 95.79$\pm$0.26 & \textbf{0.04} & \textbf{395.78} & 4.97\\
       Random Label & 99.99$\pm$0.01 & 100.00$\pm$0.00 & 100.00$\pm$0.00 & 95.44$\pm$0.13 & 0.13 & 395.43 & 3.53\\
       \midrule
       UGradSL & 90.71$\pm$4.08 & 99.90$\pm$0.16 & 99.54$\pm$0.04 & 95.64$\pm$0.25 & 2.54 & 385.79 & \textbf{0.23} \\
       UGradSL+ & 100.00$\pm$0.00 & 100.00$\pm$0.00 & 99.82$\pm$0.62 & 94.35$\pm$0.70 & 0.44 & 394.17 & 4.56\\
        \bottomrule
    \end{tabular}
    }
\end{table*}

\subsection{Random Forgetting}
We present the performance of random forgetting in CIFAR-10 and SVHN dataset in Table \ref{tab:random_app}. The observation is similar in CIFAR-100 and 20 Newsgroup given in Table \ref{tab:random}. 

\begin{table*}[t]
    \centering
    \caption{
    The experiment results of random forgetting across all the classes in CIFAR-10 and SVHN. 
    }
    \label{tab:random_app}
    \scalebox{0.85}{%
    \begin{tabular}{c|cccc|cc|c}
    \toprule
        CIFAR-10 & UA & $\mbox{MIA}_{\mbox{Score}}$ & RA & TA & Avg. Gap ($\downarrow$) & Sum. ($\uparrow$) & RTE ($\downarrow$, min) \\
    \midrule
    Retrain & 8.07$\pm$0.47 & 17.41$\pm$0.69 & 100.00$\pm$0.01 & 91.61$\pm$0.24 & - & 217.09 & 24.66 \\
        \midrule
       FT & 1.10$\pm$0.19 & 4.06$\pm$0.41 & 99.83$\pm$0.03 & 93.70$\pm$0.10 & 5.65 & 198.69 & 1.58 \\
       GA & 0.56$\pm$0.01 & 1.19$\pm$0.05 & 99.48$\pm$0.02 & 94.55$\pm$0.05 & 6.80 & 195.78 & \textbf{0.31} \\
       IU & 17.51$\pm$2.19 & 21.39$\pm$1.70 & 83.28$\pm$2.44 & 78.13$\pm$2.85 & 10.91 & 200.31 & 1.18\\
       BE & 0.00$\pm$0.00 & 0.26$\pm$0.02 & 100.00$\pm$0.00 & 95.35$\pm$0.18 & 7.24 & 195.61 & 3.17\\
       BS & 0.48$\pm$0.07 & 1.16$\pm$0.04 & 99.47$\pm$0.01 & 94.58$\pm$0.03 & 6.84 & 195.69 & 1.41 \\
       $\ell_1$-sparse & 1.21$\pm$0.38 & 4.33$\pm$0.52 & 97.39$\pm$0.31 & 95.49$\pm$0.18 & 6.61 & 198.42 & 1.82 \\
       SCRUB & 0.70$\pm$0.59 & 3.88$\pm$1.25 & 99.59$\pm$0.34 & 94.22$\pm$0.26 & 5.98 & 198.42 & 4.05\\
       Random Label & 2.80$\pm$0.37 & 18.59$\pm$3.48 & 99.97$\pm$0.01 & 94.08$\pm$0.12 & 2.24 & 198.39 & 1.98\\
       \midrule
       UGradSL & 20.77$\pm$0.75 & 35.45$\pm$2.85 & 79.83$\pm$0.75 & 73.94$\pm$0.75 & 17.15 & 209.99 & 0.45\\
       UGradSL+ & 25.13$\pm$0.49 & 37.19$\pm$2.23 & 90.77$\pm$0.20 & 84.78$\pm$0.69 & 13.23 & \textbf{237.87} & 3.07\\
       \midrule
       UGradSL (R) & 5.87$\pm$0.51 & 13.33$\pm$0.70 & 98.82$\pm$0.28 & 92.17$\pm$0.23 & \textbf{2.01} & 210.19 & 0.45 \\
       UGradSL+ (R) & 6.03$\pm$0.17 & 10.65$\pm$0.13 & 99.79$\pm$0.03 & 93.64$\pm$0.16 & 2.76 & 210.11 & 3.07 \\
    \midrule
    SVHN & UA & $\mbox{MIA}_{\mbox{Score}}$ & RA & TA & Avg. Gap ($\downarrow$) & Sum. ($\uparrow$) & RTE ($\downarrow$, min) \\
    \midrule
    Retrain &  4.95$\pm$0.03 & 15.59$\pm$0.93 & 99.99$\pm$0.01 & 95.61$\pm$0.22 & - & 216.14 & 35.65 \\
    \midrule
       FT &  0.45$\pm$0.14 & 2.30$\pm$0.04 & 99.99$\pm$0.00 & 95.78$\pm$0.01 & 4.49 & 198.52 & 2.76 \\
       GA & 0.58$\pm$0.04 & 1.13$\pm$0.02 & 99.56$\pm$0.01 & 95.62$\pm$0.01 & 4.86 & 196.89 & \textbf{0.31} \\
       FF & 0.45$\pm$0.09 & 1.30$\pm$0.12 & 99.55$\pm$0.01 & 95.49$\pm$0.03 & 4.84 & 196.79 & 6.02 \\
       BE & 0.00$\pm$0.02 & 0.02$\pm$0.17 & 100.00$\pm$0.01 & 96.14$\pm$0.02 & 5.27 & 196.16 & 1.03\\
       BS & 0.45$\pm$0.14 & 1.13$\pm$0.05 & 99.57$\pm$0.03 & 95.66$\pm$0.01 & 4.86 & 196.81 & 4.24\\
       $\ell_1$-sparse & 3.73$\pm$0.78 & 8.44$\pm$0.34 & 97.84$\pm$0.28 & 96.18$\pm$0.33 & 2.77 & 206.19 & 0.07  \\
       SCRUB & 0.35$\pm$0.20 & 4.96$\pm$0.93 & 99.94$\pm$0.02 & 95.36$\pm$0.23 & 3.88 & 200.61 & 3.24 \\
       Random Label & 8.00$\pm$0.64 & 29.40$\pm$11.92 & 98.72$\pm$0.45 & 94.04$\pm$1.10 & 4.93 & 230.16 & 1.79\\
    \midrule
       UGradSL & 6.16$\pm$0.49 & 26.35$\pm$0.40 & 94.24$\pm$0.33 & 90.55$\pm$0.27 & 5.70 & 217.3 & 0.57\\
       UGradSL+ & 25.05$\pm$4.29 & 35.42$\pm$2.13 & 92.43$\pm$5.93 & 85.36$\pm$4.80 & 14.44 & \textbf{238.26} & 4.44\\
    \midrule
       UGradSL (R) & 3.29$\pm$2.53 & 14.32$\pm$4.56 & 99.89$\pm$0.02 & 94.38$\pm$0.28 & 1.07 & 211.88 &  0.57 \\
       UGradSL+ (R) & 5.77$\pm$2.93 & 15.95$\pm$2.26 & 100.00$\pm$0.00 & 95.12$\pm$0.50 & \textbf{0.42} & 216.84 & 4.44  \\
    \bottomrule
    \end{tabular}
    }
\end{table*}

\subsection{MU with the Other Classifier}
To validate the generalization of the proposed method, we also try the other classification model. We test VGG-16 and vision transformer (ViT) on the task of random forgetting across all classes and class-wise forgetting using CIFAR-10, respectively. The results are given in Table~\ref{tab:group_vgg_app} and ~\ref{tab:vit_app}. The observation is similar in Table~\ref{tab:random} and ~\ref{tab:class_wise}, respectively.

\begin{table*}[t]
    \centering
    \caption{The experiment results of random forgetting across all the classes in CIFAR-10 using VGG-16
    }
    \label{tab:group_vgg_app}
    \scalebox{0.85}{%
    \begin{tabular}{c|cccc|cc|c}
    \toprule
        CIFAR-10 & UA & One-side MIA & RA & TA & Avg. Gap ($\downarrow$) & Sum. ($\uparrow$) & RTE ($\downarrow$, min) \\
    \midrule
    Retrain & 11.41$\pm$0.41 & 11.97$\pm$0.50 & 74.65$\pm$0.23 & 66.13$\pm$0.16 & - & 164.16 & 9.48 \\
        \midrule
       FT & 1.32$\pm$0.13 & 3.48$\pm$0.13 & 74.24$\pm$0.04 & 67.04$\pm$0.10 & 4.96 & 146.08 & 0.60\\
       GA & 1.35$\pm$0.08 & 2.18$\pm$0.66 & 73.95$\pm$0.01 & 66.88$\pm$0.01 & 5.33 & 144.36 & \textbf{0.14}\\
       IU & 1.74$\pm$0.09 & 2.16$\pm$0.61 & 73.96$\pm$0.01 & 68.88$\pm$0.00 & 5.73 & 146.74 & 0.24\\
       FF & 1.35$\pm$0.09 & 2.21$\pm$0.58 & 73.95$\pm$0.02 & 66.87$\pm$0.04 & 5.63 & 144.38 & 1.02 \\
       BE & 0.01$\pm$0.01 & 0.23$\pm$0.05 & 99.98$\pm$0.00 & 94.04$\pm$0.21 & 19.10 & 194.26 & 1.09\\
       BS & 0.01$\pm$0.01 & 0.22$\pm$0.03 & 99.98$\pm$0.01 & 94.00$\pm$0.14 & 19.09 & 194.21 & 3.17\\
       $\ell_1$-sparse & 1.27$\pm$1.13 & 3.60$\pm$2.41 & 98.97$\pm$1.13 & 92.18$\pm$1.46 & 17.22 & \textbf{196.02} & 0.08\\ 
       SCRUB & 61.16$\pm$50.89 & 44.65$\pm$43.31 & 39.26$\pm$50.57 & 36.95$\pm$46.68 & 36.75 & 182.02 & 0.91\\ 
       \midrule
       UGradSL & 13.45$\pm$0.63 & 11.77$\pm$0.54 & 65.05$\pm$0.48 & 58.52$\pm$0.38 & \textbf{4.86} & 148.79 & 0.19\\
       UGradSL+ & 12.41$\pm$0.32 & 14.96$\pm$0.52 & 65.90$\pm$0.52 & 58.58$\pm$0.35 & 5.13 & 151.85 & 1.08\\
    \bottomrule
    \end{tabular}
    }
\end{table*}

\begin{table*}[t]
    \centering
    \caption{
    The experiment results of class-wise forgetting in CIFAR-10 using ViT. 
    }
    \label{tab:vit_app}
    \scalebox{0.9}{%
    \begin{tabular}{c|cccc|cc|c}
    \toprule
        CIFAR-10 & UA & One-side MIA & RA & TA & Avg. Gap ($\downarrow$) & Sum. ($\uparrow$) & RTE ($\downarrow$, min) \\
    \midrule
    Retrain & 100.00$\pm$0.00 & 100.00$\pm$0.00 & 61.41$\pm$0.81 & 58.94$\pm$1.09 & - & 320.35 & 189.08 \\
        \midrule
       FT & 3.97$\pm$0.87 & 7.60$\pm$1.76 & 98.29$\pm$0.05 & 80.44$\pm$0.22 & 61.70 & 190.3 & 2.99 \\
       GA & 33.77$\pm$6.36 & 40.47$\pm$6.63 & 89.47$\pm$4.21 & 71.65$\pm$2.79 & 41.63 & 235.36 & 0.32 \\
       IU & 1.74$\pm$0.09 & 2.16$\pm$0.61 & 73.96$\pm$0.01 & 68.88$\pm$0.00 & 54.65 & 146.74 & 0.24\\
       FF & 1.35$\pm$0.09 & 2.21$\pm$0.58 & 73.95$\pm$0.02 & 66.87$\pm$0.04 & 54.23 & 144.38 & 1.02 \\
       BE & 85.56$\pm$3.07 & 99.98$\pm$0.02 & 99.55$\pm$0.01 & 95.53$\pm$0.07 & 22.30 & 380.62 & 3.17\\
       \midrule
       UGradSL & 68.11$\pm$11.03 & 73.84$\pm$9.58 & 84.11$\pm$2.70 & 68.33$\pm$1.69 & 22.54 & \textbf{294.39} & \textbf{0.22}\\
       UGradSL+ & 99.99$\pm$0.01 & 99.99$\pm$0.02 & 94.46$\pm$1.06 & 77.26$\pm$1.19 & \textbf{12.85} & 371.7 & 5.86\\ 
    \bottomrule
    \end{tabular}
    }
\end{table*}

\subsection{The Effect of the Smooth Rate}
\begin{figure}[ht]
    \centering
    \includegraphics[width=\textwidth]{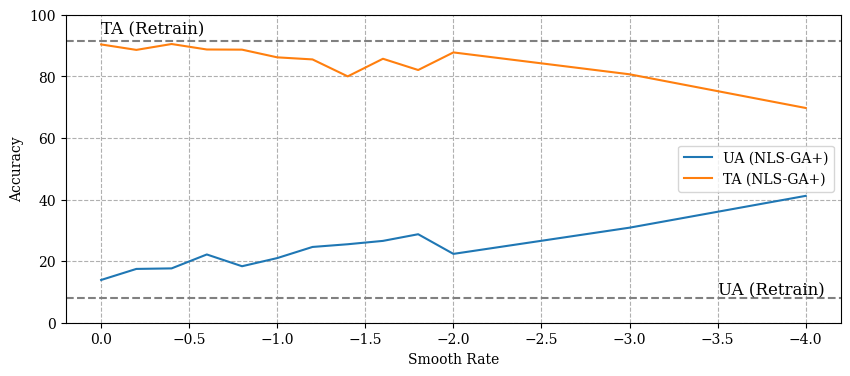}
    \caption{The relationship between the performance and smooth rate in random forgetting across all classes using CIFAR-10. The gray dash line stands for the performance of retrain. Our methods can improve the unlearning accuracy (UA) without significant drop of testing accuracy (TA).}
    \label{fig:smooth_rate}
\end{figure}

We also investigate the relationship between the performance and the smooth rate $\alpha$. 
We select UGradSL+ using the random forgetting across all classes in CIFAR-10. 
The results are given in Figure~\ref{fig:smooth_rate}. 
It should be noted that for the completely unlearning, \textbf{the hyper-parameter tuning is acceptable using UGradSL and UGradSL+ because the most important metrics are from $D_f \in D_{tr}$.}
For the retraining-based method, we use the validation set to select the hyper-parameters.
We do not use any extra information from the testing dataset and retrained model which should not be accessed in practice.
Our method can improve the unlearning accuracy (UA) without significant drop of testing accuracy (TA).

\subsection{Streisand effect}
\label{appendix:streisand_effect}
\begin{figure}[ht]
    \centering
    \includegraphics[width=\textwidth]{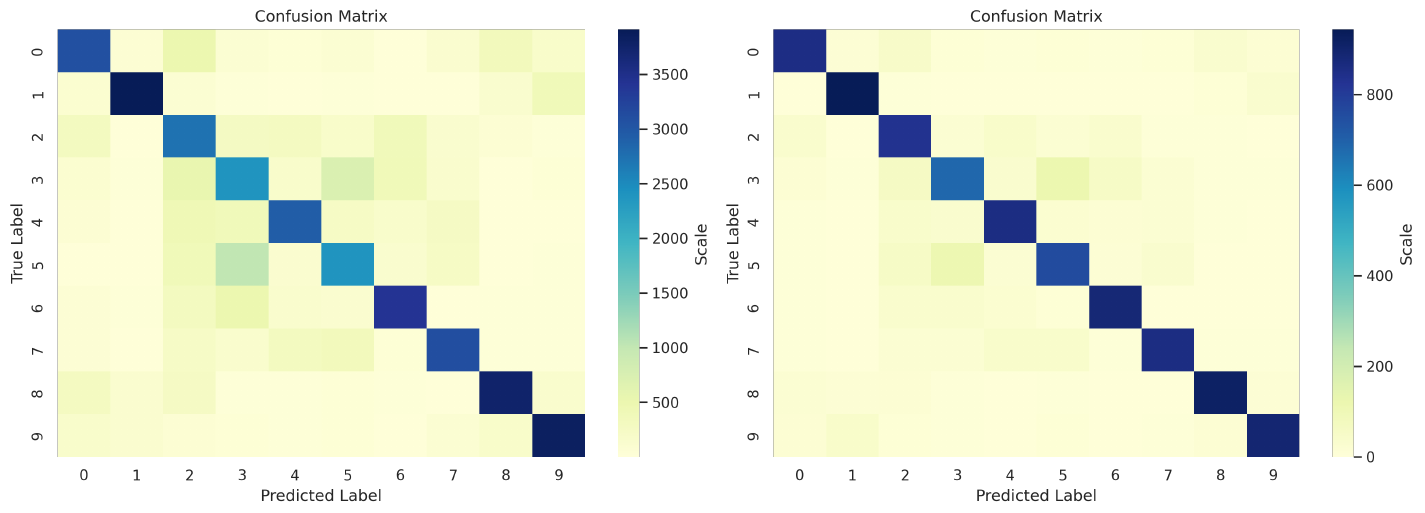}
    \caption{The confusion matrix of testing set and forgetting set $D_f$ using our method on CIFAR-10 with random forgetting across all the classes. There is no big difference between the prediction distribution. Our method will not make $D_f$ more distinguishable.}
    \label{fig:streisand}
\end{figure}

From the perspective of security, it is important to make the predicted distributions are almost the same from the forgetting set $D_f$ and the testing set $D_{te}$, which is called Streisand effect. We investigate this effect in the \textit{random forgetting} on CIFAR-10 by plotting confusion matrix as shown in Figure \ref{fig:streisand}. It can be found that our method will not lead to the extra hint of $D_f$.

\subsection{Gradient Analysis}
\label{appendix:grad_analysis}
\begin{figure}[ht]
    \centering
    \includegraphics[width=\textwidth]{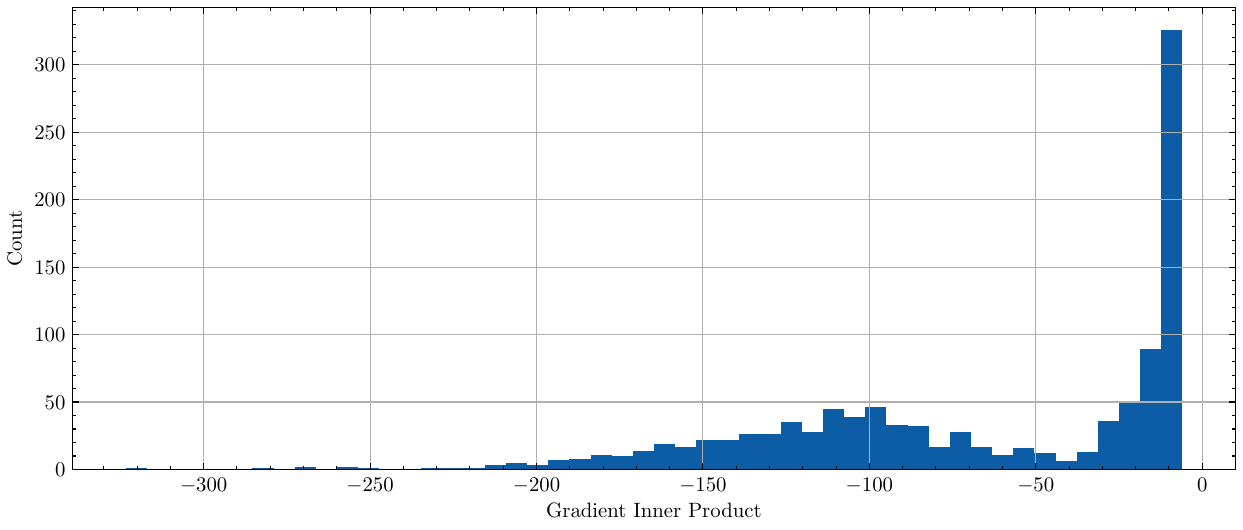}
    \caption{The distribution of $\langle \Delta \btheta_{r} - \Delta \btheta_{f}, \Delta \btheta_n-\Delta \btheta_{f}\rangle$ on CelebA dataset.}
    \label{fig:grad_distribution}
\end{figure}

As mentioned in Section \ref{sec:nls_MU}, $\langle \Delta \btheta_{r} - \Delta \btheta_{f}, \Delta \btheta_n-\Delta \btheta_{f}\rangle \le 0$ always holds practically. We practically check the results on CelebA dataset. The distribution of $\langle \Delta \btheta_{r} - \Delta \btheta_{f}, \Delta \btheta_n-\Delta \btheta_{f}\rangle$ is shown in Figure \ref{fig:grad_distribution}, which is with our assumption.

\subsection{The difference between UGradSL and UGradSL+}
\label{appendix:difference}
Although UGradSL and UGradSL+ look similar, the intuition of these two method is totally different because of the difference between FT and GA. We conducted experiments to illustrate the difference between GA and FT as well as UGradSL and UGradSL+. The results are given in Table \ref{tab:ugradsl_ugradsl_comp}.
The dataset and forgetting paradigm is CIFAR-10 random forgetting. It can be found that the difference becomes much larger when the number of epochs is over 8. When the number of epochs is 10, the model is useless because TA is less than 10\%. We also report the performance of UGradSL and UGradSL+ in different epochs. For UGradSL, when the epochs are over 14, the model cannot be used at all.
For UGradSL+, the algorithm is much more stable, showing the very good adaptive capability.

\begin{table}[]
    \centering
    \caption{Baselines with retrain as standard.}
    \scalebox{0.7}{%
    \begin{tabular}{c|cccc|cc||cccc|cc}
    \toprule
    &  \multicolumn{6}{c}{CIFAR-10} & \multicolumn{6}{c}{CIFAR-100} \\ 
    \midrule
     & UA & $\mbox{MIA}_{\mbox{Score}}$ & RA & TA & Avg. Gap ($\downarrow$) & Sum ($\uparrow$) & UA & $\mbox{MIA}_{\mbox{Score}}$ & RA & TA & Avg. Gap ($\downarrow$) & Sum ($\uparrow$) \\
    \midrule
    FT (R) & 8.19 & 9.01 & 92.56 & 98.27 & 5.60& 208.03 & 19.86 & 26.66 & 93.25 & 67.44	& 11.56 & 207.21\\
    GA (R) & 8.01 & 10.35 & 95.20 & 100 & 5.07 & 213.56 & 6.03 &  6.53 & 99.97 & 78.22 & 19.53 & 190.75\\
    IU (R) & 10.17 & 16.36 & 92.67 & 99.02 & 4.47 & 218.22 & 8.58 & 17.03 & 91.67 & 66 &  17.55 & 183.28\\
    RL (R) &  8.80 & 17.59 & 99.98 & 98.97 & 2.07 & 225.31 & 40.06 & 93.43& 99.94 & 70.91& 12.74& 304.34\\
    BE (R) & 0.00 & 10.26 & 99.35 & 92.67 & 4.23 & 202.28 & 1.45	& 1.45 & 99.97 & 78.26	& 21.96 & 181.13\\
    BS (R) & 0.70& 9.93&	94.03&	91.27&	5.29&	195.93&	2.30&	10.92&	98.18&	70.06&	18.00&	181.46 \\
    $\ell_1$-sparse (R) & 5.69	&12.08&	98.17&	96.80	&3.69&	212.74 & 8.26&	18.96&	97.22&	71.63&	14.90&	196.07 \\
    UGradSL &20.77	&35.45&	79.83	&73.94&	17.15&	209.99 &15.10&	34.67	&86.69	&59.25&	14.44&	195.71 \\
    UGradSL+ & 25.13&	37.19	&90.77&	84.78	&13.23&	\textbf{237.87} & 63.89&	71.51&	92.25&	61.09&	17.40&	\textbf{288.74}\\
    UGradSL (R) & 5.87&	13.33	&98.82&	92.17&	\textbf{2.01} &	210.19&18.36&	40.71	&98.38&	68.23	&6.95	&207.95 \\
    UGradSL+ (R) & 6.03	&10.65&	99.79	&93.64&	2.76&	210.11&21.69&	49.47	&99.87	&73.60	& \textbf{3.75}	&244.63 \\
    \midrule
    Retrain & 8.07 & 17.41 & 100.00 & 91.61 & 	- & 217.09 & 29.47 & 53.50 & 99.98 & 70.51 & - &253.46\\
    \bottomrule
    \end{tabular}
    }
    \label{tab:faircom}
\end{table}

\subsection{Other baseline compared with Retraining}

\begin{table}[]
    \centering
    \caption{The difference between GA and FT as well as UGradSL and UGradSL+ on CIFAR-10 regarding the number of epochs. The forgetting paradigm is random forgetting. }
    \scalebox{0.8}{%
    \begin{tabular}{c|cccc|cc||cccc|cc}
    \toprule
    &  \multicolumn{6}{c}{Gradient Ascent} & \multicolumn{6}{c}{Fine-tuning} \\ 
    \midrule
    Epoch & UA & $\mbox{MIA}_{\mbox{Score}}$ & RA & TA & Avg. Gap & Sum & UA & $\mbox{MIA}_{\mbox{Score}}$ & RA & TA & Avg. Gap & Sum \\
    \midrule
    5 & 0 & 0.32 & 95.31 & 100 & 3.98 & 195.63 & 0.04 & 0.34 & 95.13 & 99.99 & 3.96 & 195.50\\
    6 & 0 & 0.40 & 95.34 & 100 & 3.96 & 195.74 & - & - & - & - & - & - \\
    7 & 0.82 & 2.22 & 93.24 & 99.26 & 3.95 & 195.54 & - & - & - & - & - & -\\
    8 & 3.44 & 4.78 & 90.80 & 96.18 & 4.03 & 195.20 & - & - & - & - & - & -\\
    9 & 10.34 & 12.76 & 83.42 & 89.00 & 7.44 & 195.52 & - & - & - & - & - & -\\
    10 & 76.26 & 72.22 & 6.49 & 24.24 & 74.21 & 179.21 & 0.04 & 0.24 & 94.97 & 99.99 & 4.02 & 195.24 \\
    15 & - & - & - & - & - &  - & 0.02 & 0.80 & 94.68 &  99.96 & 3.97 & 195.46\\	
    \midrule
    &  \multicolumn{6}{c}{UGradSL} & \multicolumn{6}{c}{UGradSL+} \\ 
    \midrule
    Epoch & UA & $\mbox{MIA}_{\mbox{Score}}$ & RA & TA & Avg. Gap & Sum & UA & $\mbox{MIA}_{\mbox{Score}}$ & RA & TA & Avg. Gap & Sum \\
    \midrule
    10	& 14.98	& 33.22	& 77.18  & 84.07 & 16.51 & 209.45 & 6.26 & 14.10 & 93.39 & 99.62 & 1.33 & 213.37 \\
    11	& 24.26 & 34.38 & 68.22 & 75.06	& 23.61 & 201.92 & 6.52 & 11.66 & 93.04 & 99.37 & 1.21 & 210.59 \\
    12 & 28.70 & 24.62 & 68.17 & 74.39 & 22.46 & 195.88 & 21.46	& 27.38 & 89.41 & 97.07 & 10.36 & 235.32 \\
    13 & 38.46 & 72.90 & 61.78 & 64.72 & 40.99 & 237.86 & 29.48 & 31.92 & 87.74 & 94.93 & 14.46 & 244.07 \\
    14 & 99.86 & 86.74 & 0.45 & 0.20 & 91.26 & 187.25 & 31.62 & 32.68 & 86.53 & 93.36 & 15.88 & 244.19 \\
    \midrule
    Retrain & 4.5 & 11.62 & 95.21 & 100 & - & 211.33 & 4.5 & 11.62 & 95.21 & 100 & - & 211.33 \\
    \bottomrule
    \end{tabular}
    }
    \label{tab:ugradsl_ugradsl_comp}
\end{table}



\begin{figure}[ht]
	\centering
	\begin{subfigure}{0.45\linewidth}
		\includegraphics[width=\linewidth]{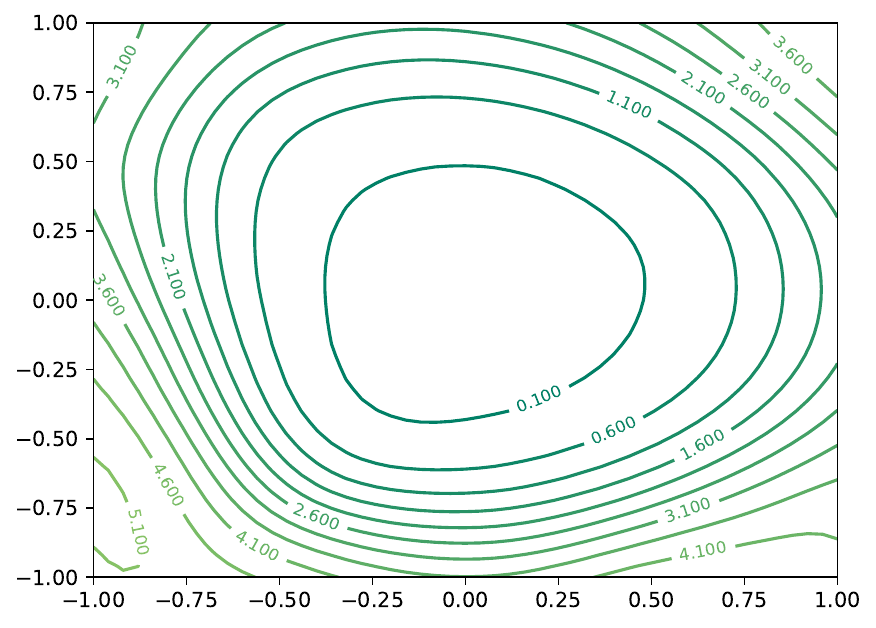}
		\label{fig:subfigA}
	\end{subfigure}
	\begin{subfigure}{0.45\linewidth}
		\includegraphics[width=\linewidth]{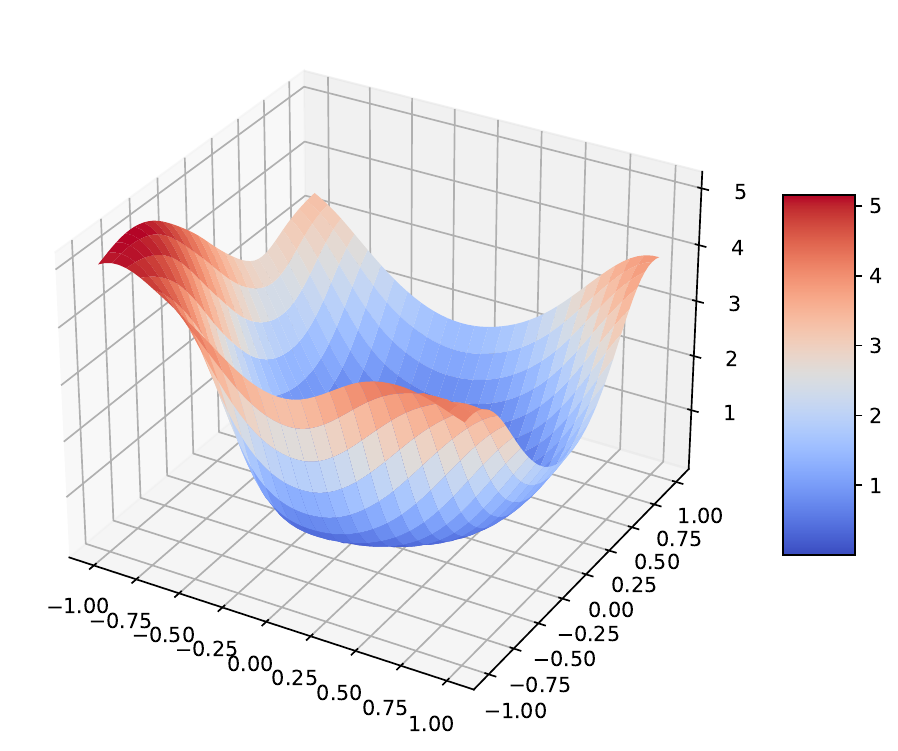}
		\label{fig:subfigB}
	\end{subfigure}
	\caption{The loss land scape of $\btheta_r$ on CIFAR-10 and the model is ResNet-18.}
	\label{fig:loss_landscape}
\end{figure}

\begin{sidewaysfigure}[ht]
\includegraphics[width=\textwidth,]{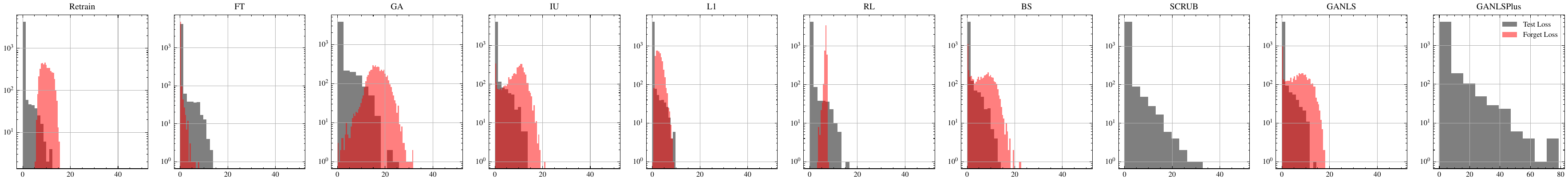}
\includegraphics[width=\textwidth,]{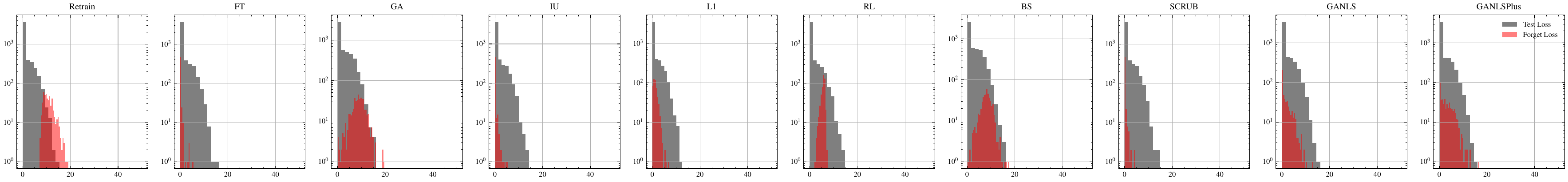}
\includegraphics[width=\textwidth,]{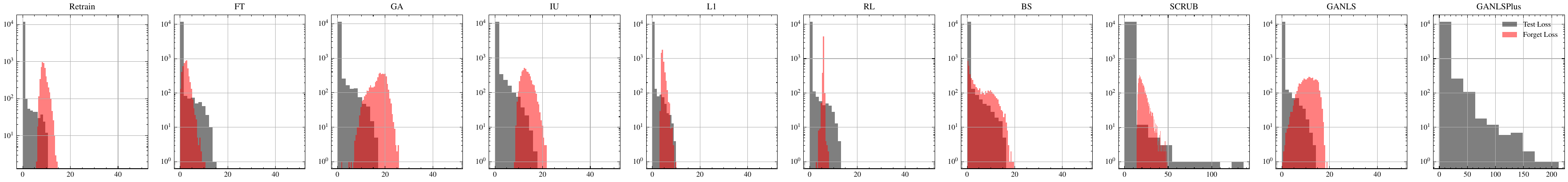}
\includegraphics[width=\textwidth,]{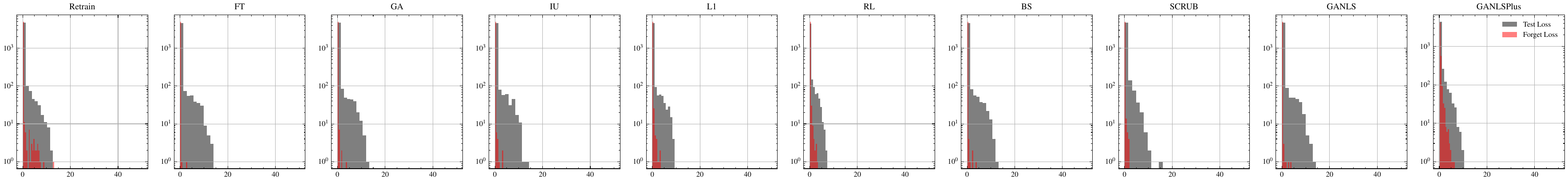}
\includegraphics[width=\textwidth,]{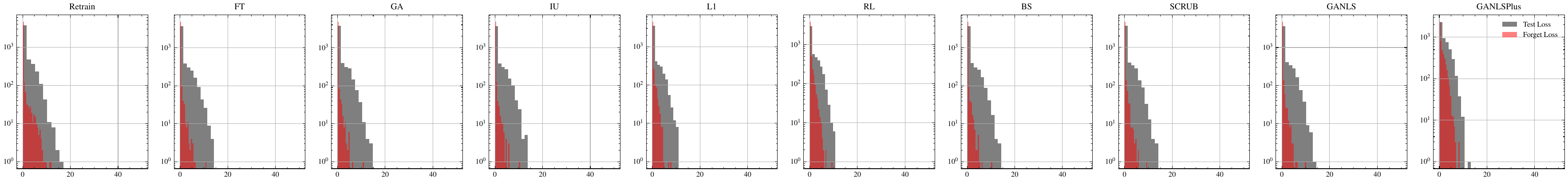}
\includegraphics[width=\textwidth,]{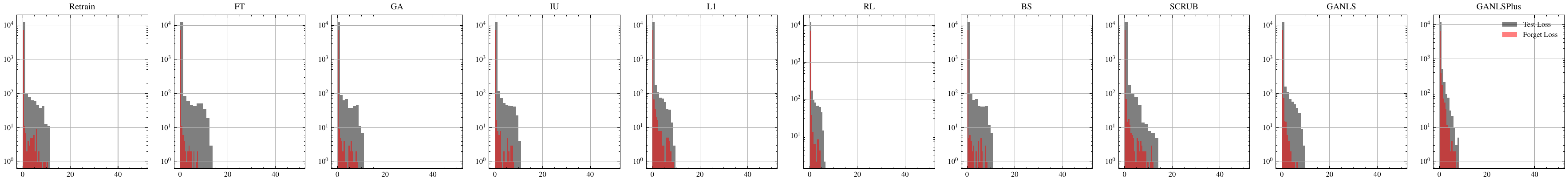}
\caption{The distributions of the cross-entropy losses for the forget and test instances from the unlearned models. The y-axis is in log scale for better visualization. From the first to the last figure, they are random forgetting on CIFAR-10, CIFAR-100, SVHN and class-wise forgetting on CIFAR-10, CIFAR-100, SVHN.}
\label{fig:loss_dist}
\end{sidewaysfigure}

\end{document}